# Learning to Optimize: Joint Routing and Flow Allocation on Sparse Non-Euclidean Networks

Haomiao Sun, Fang He*, Congyuan Ji

Department of Industrial Engineering, Tsinghua University, Beijing 100084, P.R. China, shm21@mails.tsinghua.edu.cn, fanghe@tsinghua.edu.cn (corresponding author), jcy20@mails.tsinghua.edu.cn

Xindi Tang

School of Management Science and Engineering, Central University of Finance and Economics, Beijing 100081, P.R. China, tangxd@cufe.edu.cn

**Abstract.** We study an integrated pickup-and-delivery problem on sparse, non-Euclidean networks that jointly optimizes cyclic routing, cargo flow allocation, and cross-cycle service. The tight coupling of these operational constraints creates a complex discrete-continuous decision space with highly restricted feasible regions. To overcome these computational challenges, we propose Double-Channel Graph Attention (DCGA), an end-to-end reinforcement learning framework. DCGA isolates network reachability and demand-service logic into separate graph channels and constructs valid routes using a simulator-coupled, constraint-informed decoder. Experiments on LinerLib benchmarks demonstrate that DCGA achieves seconds-level inference and delivers state-of-the-art solution quality on instances beyond a specific scale, with its advantage over existing baselines widening significantly as problem size increases. Supported by extensive stability and ablation analyses, our results demonstrate that this structure-aware learning approach provides an effective, low-latency engine for realistic routing-and-flow optimization.

**Funding:** This research was partially supported by the National Natural Science Foundation of China [Grant Nos. 72371141 and 72301305].

## 1. Introduction

The operational efficiency of modern large-scale transportation systems, particularly in domains such as container liner shipping, fundamentally hinges on the tightly coupled optimization of cyclic routing and cargo flow allocation. Unlike the fully connected, idealized planar coordinate systems frequently studied in academic literature, real-world industrial networks are inherently sparse, directed, and strictly non-Euclidean. Within these networks, physical nodes (e.g., hub ports) often play multiple simultaneous roles as origins, destinations, and transshipment hubs. Furthermore, because vessels operate on continuous loops, they need to execute cross-cycle service—where cargo picked up in a previous voyage is delivered within the current reference cycle—under rigorous capacity limitations and soft due-date commitment. To formally capture these intertwined complexities, we investigate the integrated Non-Euclidean Pickup-and-Delivery Problem with Cross-Cycle Service (NE-PDP-CCS).

While traditional exact methods, such as mixed-integer programming (MIP) and dynamic programming (DP), and advanced heuristics/metaheuristics like Lin-Kernighan-Helsgaun heuristic (LKH3) (Helsgaun 2017) and adaptive large neighborhood search (ALNS) (Dragomir et al. 2022) excel in classical vehicle routing, they struggle with the NE-PDP-CCS. Exact methods suffer the curse of dimensionality, precluding seconds-level replanning, whereas heuristics fail to consistently find high-quality joint solutions under strict time limits. Consequently, the industry urgently requires a low-latency, high-quality joint decision algorithm framework capable of satisfying the operational realities of dynamic rolling replanning. Deep reinforcement learning (DRL) offers a promising end-to-end alternative, yet existing neural solvers, such as policy optimization with multiple optima (POMO) (Kwon et al. 2020) and neural collaborative search (NCS) (Kong et al. 2024), are primarily tailored to fully connected Euclidean planes. Relying on the assumption of direct straight-line connectivity, these models frequently generate infeasible routes and fail to capture meaningful topological representations when applied to sparse, asymmetric real-world networks. Furthermore, conventional DRL frameworks structurally lack the capacity to handle cross-cycle service, a fundamental requirement for cyclic logistics.

To bridge the gap between idealized academic benchmarks and complex industrial decisions, we formally propose the NE-PDP-CCS abstraction. This formulation abandons the oversimplified coordinate-based geometries to faithfully reflect the sparse, non-Euclidean nature of practical logistics. The NE-PDP-CCS is well aligned with real-world industrial scenarios, most notably liner shipping networks. In such domains, maritime routes naturally operate on predefined, non-Euclidean sea lanes and require continuous, cross-cycle cyclic scheduling to balance massive cargo flows against limited vessel capacities and varying delivery deadlines.

To overcome these profound complexities, we propose the Double-Channel Graph Attention (DCGA) framework. DCGA's double-channel architecture isolates physical reachability from supply-demand logic to prevent representational collapse. Furthermore, it elegantly accommodates cross-cycle service by utilizing an origin-destination (OD)-indexed logical action space and a delivery-side bias, thereby bypassing rigid precedence constraints. Guided by a deterministic simulator and constraint-aware masks, DCGA effectively learns constrained exploration in sparse spaces, seamlessly aligning routing feasibility with flow-allocation profitability.

The superior optimization capability of this framework is most prominently demonstrated through extensive evaluations on the classic maritime public dataset, LinerLib (Brouer et al. 2014). Particularly on large-scale instances, DCGA comprehensively dominates the existing state-of-the-art mathematical optimization benchmark, advanced metaheuristics, and contemporary neural routing solvers in both solution quality and computational efficiency. Crucially, our model maintains a definitive performance edge even when competing against an aggressively adapted NCS baseline that has been specifically enhanced with a graph encoder. By consistently delivering superior, high-quality solutions with seconds-level inference speeds, DCGA exhibits highly robust generalization, proving its immense potential as an excellent, low-latency decision engine for large-scale industrial applications.

### 1.1. Related Work

A substantial body of research has investigated the pickup-and-delivery problem (PDP) and its variants through diverse formulations and solution approaches (Zhou et al. 2025). These studies are typically tailored to specific operational contexts and therefore focus on different subsets of problem features. By contrast, we consider a broader class of integrated PDPs characterized by several rarely combined structural features: sparse non-Euclidean travel costs and times, capacity-coupled routing and fulfillment, multi-task physical nodes, optional fulfillment, soft due-date commitment, and cross-cycle service.

On the *traditional optimization* side, PDP variants have commonly been formulated as MIP models or solved through exact and DP-based methods (Ruland and Rodin 1997, Bianco et al. 1994, Shelbourne et al. 2017, Li et al. 2023). These methods provide exact optimization foundations and can certify or closely approach optimality on small and structured instances, but their computational burden often grows rapidly with instance size. For example, Bianco et al. (1994) adopted DP for related routing problems with precedence constraints, exploiting their temporal decomposition structure. Li et al. (2023) developed a branch-price-and-cut algorithm for a split pickup-and-delivery routing problem in which routing and pickup/delivery quantities are optimized jointly.

*Heuristics and metaheuristics* remain highly competitive for practical-scale routing problems (Liu et al. 2023, Lin 1965, Kirkpatrick et al. 1983, Jung and Haghani 2000, Huang and Ting 2010, Dragomir et al. 2022, Helsgaun 2000, 2017). These methods usually combine fast construction rules with local search or population-based improvement, and they have been successfully adapted to many PDP and vehicle routing problem (VRP) variants. Among them, ALNS is widely used as a strong metaheuristic for rich VRP/PDP variants because its operators can be adapted to complex operational constraints (Dragomir et al. 2022, Ropke and Pisinger 2006). Lin-Kernighan-Helsgaun-type methods, especially LKH3, are also commonly regarded as among the strongest classical baselines for constrained TSP, VRP, and PDP variants (Helsgaun 2000, 2017).

On the *learning-based* side, DRL and neural combinatorial optimization (NCO) have attracted increasing attention because they enable end-to-end policy learning and fast test-time inference (Lee et al. 2024, Nazari et al. 2018, Vaswani et al. 2017, Kool et al. 2019, Zhou et al. 2025). Attention model (AM)-style policies have been extended to richer structures, including soft time windows, pickup-and-delivery asymmetry, late-penalty PDPs, and symmetry-aware pickup-and-delivery routing (Zhang et al. 2020, 2022, Li et al. 2024). In the improvement line, Kong et al. (2024) reported that NCS/N2S achieved state-of-the-art performance among neural methods on canonical PDP, combining learned construction with collaborative improvement mechanisms.

Despite these advances, key gaps remain. Table 1 summarizes representative studies according to the structural features. The comparison highlights two main limitations. First, most existing methods focus on a limited combination of operational features, whereas practical routing-and-flow problems often require most, if not all, of these features to be considered jointly. Second, existing learning-based methods are mainly developed for Euclidean and fully connected settings, and they pay limited attention to richer operational variants such as sparse reachability, asymmetric travel attributes, and cyclic service logic. To the best of our knowledge, no existing end-to-end learning framework solves this integrated class of non-Euclidean PDPs in a unified model.

**Table 1 Representative related work with the structural features considered in the NE-PDP-CCS.**

| Category | Representative work | PDP | NE | C | TW | DC | MD | SN | MN | OF | CCS |
|---|---|---|---|---|---|---|---|---|---|---|---|
| Optimization | Branch-and-cut (Ruland and Rodin 1997) | ✓ | ✓ | | | | | | | | |
| | Branch-price-and-cut (Li et al. 2023) | ✓ | ✓ | ✓ | | | ✓ | | | | |
| | DP (Bianco et al. 1994) | ✓ | ✓ | | | | | | | | |
| Heuristic & Metaheuristics | GA (Jung and Haghani 2000) | ✓ | ✓ | ✓ | ✓ | | | | | | |
| | SA (Li and Lim 2001) | ✓ | ✓ | ✓ | ✓ | | | | | | |
| | ACO (Huang and Ting 2010) | ✓ | ✓ | ✓ | ✓ | | | | | | |
| | LKH3 (Helsgaun 2017) | ✓ | ✓ | ✓ | | | | | | | |
| | ALNS (Dragomir et al. 2022, Ropke and Pisinger 2006) | ✓ | ✓ | ✓ | ✓ | | ✓ | | | | |
| Learning-based | RL-BS (Nazari et al. 2018) | | | ✓ | | | | | | | |
| | AM / MAAM (Kool et al. 2019, Zhang et al. 2020) | | | ✓ | ✓ | | | | | | |
| | POMO (Kwon et al. 2020) | | | ✓ | | | | | | | |
| | Joint T-RL / PD-SNO (Zhang et al. 2022, Li et al. 2024) | ✓ | | ✓ | ✓ | | | | | | |
| | NCS / N2S (Kong et al. 2024) | ✓ | | | | | | | | | |
| Ours | DCGA (this paper) | ✓ | ✓ | ✓ | | ✓ | ✓ | ✓ | ✓ | ✓ | ✓ |

*Notes.* NE = Non-Euclidean travel cost/time; C = Capacity; TW = Time window; DC = Due-date commitment; MD = Maximum duration; SN = Sparse network; MN = Multi-task nodes; OF = Optional fulfillment; CCS = Cross-cycle service. For rows containing multiple methods, a checkmark indicates that at least one cited method considers the corresponding feature.

### 1.2. Contributions

The main contributions of this paper are summarized as follows.

First, we formalize the NE-PDP-CCS, systematically bridging the fundamental gap between idealized neural combinatorial optimization benchmarks and the constrained realities of industrial operations. Moving beyond the stylized, fully connected Euclidean planes prevalent in current learning-based literature, our abstraction establishes a rigorous environment characterized by sparse reachability, asymmetric travel attributes, multi-task physical hubs, and cross-cycle service dynamics. By seamlessly integrating these interrelated features, our formulation provides a vital, operationally faithful testbed for deploying learning-based solvers in complex maritime and logistics systems.

Second, we develop DCGA, an end-to-end reinforcement learning solver explicitly tailored to the structural complexities of the NE-PDP-CCS. The core methodological breakthrough lies in its double-channel architecture, which physically isolates sparse network reachability (the network channel) from supply-demand service logic (the demand channel) during feature extraction to prevent representational collapse. To systematically translate these dual representations into strictly feasible decisions, we engineer a dynamic state-conditioned decoder fortified by a composite operational masking mechanism. Rather than relying on computationally expensive post-hoc repairs, our architecture preemptively neutralizes the challenges of sparse connectivity, dynamically evolving capacity, and multi-task physical hubs via step-wise hard masks. Furthermore, by synergizing an OD-indexed logical action space with a regularizing delivery-side bias, the decoder elegantly accommodates optional fulfillment and the counter-intuitive temporal inversions of cross-cycle deliveries without requiring rigid precedence constraints. Continuously evaluated by a deterministic flow simulator, this constraint-aware mechanism empowers the policy to execute valid, profit-aligned exploration across a tightly restricted combinatorial space.

Third, we establish a new performance benchmark for industrial logistics through extensive computational evaluations on the classic maritime public dataset, LinerLib. Particularly on large-scale instances, DCGA comprehensively dominates state-of-the-art mathematical optimization benchmark (time-limited Gurobi MIP), advanced metaheuristics (LKH3, ALNS, GA, SA), and aggressively adapted neural routing solvers (NCS) in both solution quality and computational efficiency. Crucially, DCGA maintains a consistent performance advantage even when competing against an aggressively adapted NCS baseline specifically enhanced with our graph encoder. This direct comparison rigorously demonstrates that our performance breakthroughs are driven not merely by improved spatial feature extraction, but by the holistic integration of our double-channel architecture and constraint-guided decoding mechanism. As instance sizes scale, DCGA's performance advantage widens significantly, consistently delivering superior solutions with seconds-level inference speeds. Supported by comprehensive ablation and stability analyses, our results demonstrate that DCGA exhibits exceptional robustness against tested combinatorial variations and demand perturbations, proving its immense potential as an excellent, low-latency decision engine for large-scale applications.

The remainder of this paper is organized as follows. Sec. 2 formalizes the problem and presents the objective and constraints. Sec. 3 details the tailored neural network architecture, training strategy, and implementation. Sec. 4 illustrates the experimental setup, results, and ablations. Sec. 5 concludes and discusses future research directions.

## 2. Problem Formulation

For the NE-PDP-CCS, the planner determines a feasible service loop, the OD requests to serve, and the corresponding cargo flow, balancing fulfillment revenue, travel cost, unmet-demand penalty, and soft due-date commitment penalty. The formulation must capture three coupled features: routing feasibility is restricted by the sparse directed physical network; routing and cargo flow allocation interact through capacity and optional fulfillment; and cyclic service allows cross-cycle effects, where a delivery may appear before its paired pickup within the reference cycle.

Let $\mathcal{K}$ denote the set of physical nodes and $\mathcal{R} = \{1, \ldots, N\}$ the set of OD requests. Because a physical node may host multiple pickup-and-delivery tasks, and a visit to that node does not necessarily execute all local services, we use an OD-indexed logical node set. Each request $r \in \mathcal{R}$ induces a pickup node $p_r = r$ and a delivery node $d_r = r + N$, and we define

$$\mathcal{V} = \mathcal{V}^P \cup \mathcal{V}^D, \qquad \mathcal{V}^P = \{1, \ldots, N\}, \qquad \mathcal{V}^D = \{N+1, \ldots, 2N\}. \tag{1}$$

Logical nodes represent request-specific service roles rather than distinct physical locations. The mapping $\phi : \mathcal{V} \to \mathcal{K}$ maps each logical node to its physical node, with $\phi(p_r)$ and $\phi(d_r)$ corresponding to the origin and destination of request $r$. Travel time and cost are evaluated on the physical network through this mapping, while tasks associated with the same physical port may still occupy different service positions and timestamps.

Fig. 1 illustrates the main idea of the formulation. The upper panel shows that the soft due-date commitment is defined on the *relative* time elapsed between pickup and delivery, rather than as an absolute time window. The lower panel shows the corresponding sequence-space representation:

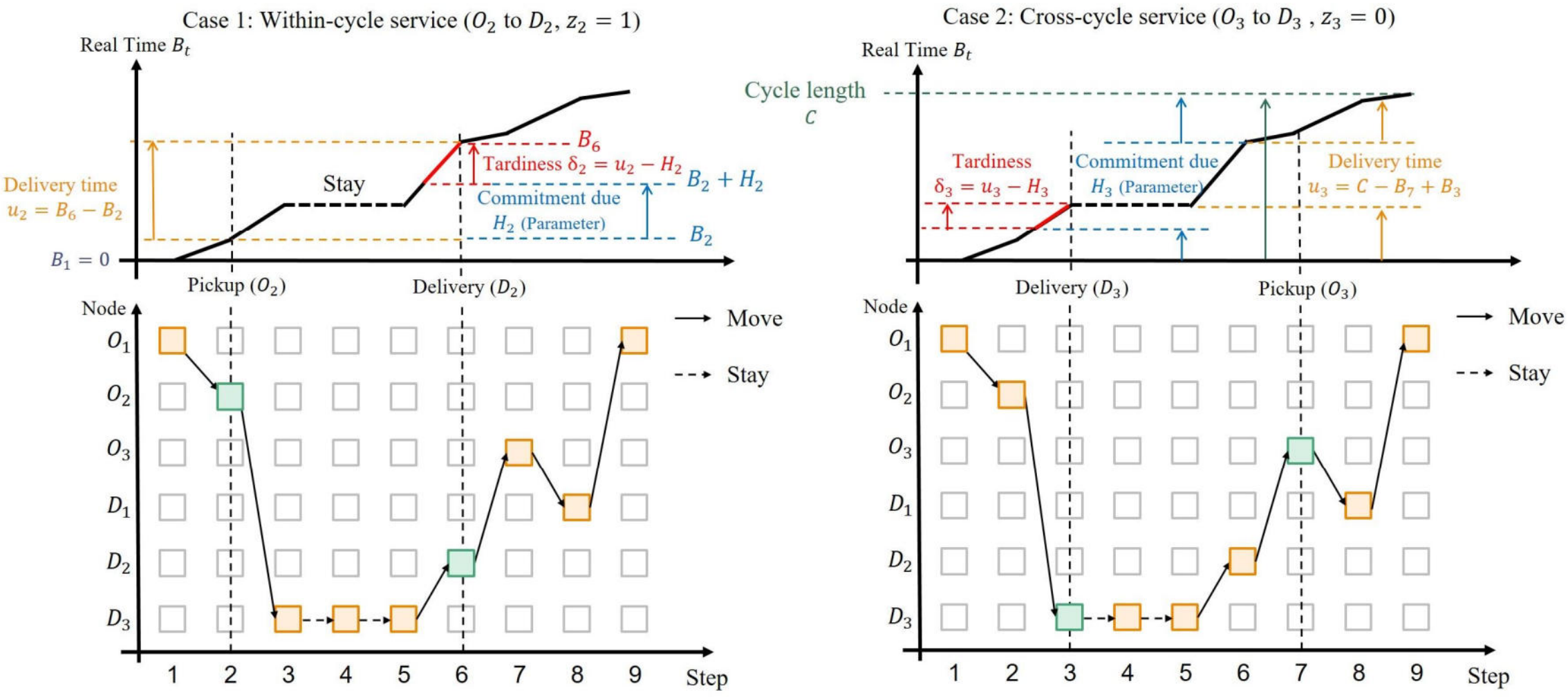

**Figure 1** **Illustration of the sequence-space representation and the soft due-date commitment horizon.**

each row corresponds to a logical service node, each column corresponds to one service position in the cyclic route, and a feasible route is represented by a path through this matrix. Routing decisions are defined on a sequence-space network with $|\mathcal{V}|$ rows and $\bar{T}$ columns, where $\bar{T}$ is a route-length upper bound. The binary variable $y_{i,t}$ indicates whether logical node $i$ is selected at $t$, and $x_{ij,t}$ indicates whether the route moves from node $i$ at position $t$ to node $j$ at position $t+1$, with column $\bar{T}$ connected back to the first column to form a cycle. If the corresponding physical movement $(\phi(i), \phi(j))$ is infeasible, we fix $x_{ij,t} = 0$. Let $\mathcal{T} = \{1, \ldots, \bar{T}\}$ be the set of service positions. For physical nodes $k, l \in \mathcal{K}$, $c_{kl}$ and $\tau_{kl}$ denote travel cost and time. For request $r$, $q_r, \rho_r, \pi_r, \varphi_r, H_r$ denote demand quantity, unit fulfillment revenue, unit unmet-demand penalty, unit tardiness penalty, and relative due-date commitment horizon, respectively. Finally, let $Q$ be the vehicle capacity, $C_{\max}$ the maximum cycle length, $\mathcal{I}$ the set of infeasible logical transitions, and $M$ a sufficiently large constant.

Model (2)-(27) maximizes total revenue minus travel cost, soft due-date commitment penalty, and unmet-demand penalty. In objective (2), $w_r$ and $\delta_r$ denote the fulfilled quantity and tardiness of request $r$, respectively. $\rho_r w_r$ is the corresponding revenue, $c_{\phi(i)\phi(j)} x_{ij,t}$ is the travel cost, $\varphi_r w_r \delta_r$ is the soft due-date commitment penalty, and $\pi_r(q_r - w_r)$ penalizes unmet demand.

$$\max \quad \sum_{r \in \mathcal{R}} \rho_r w_r - \sum_{t \in \mathcal{T}} \sum_{i \in \mathcal{V}} \sum_{j \in \mathcal{V}} c_{\phi(i)\phi(j)}\, x_{ij,t} - \sum_{r \in \mathcal{R}} \varphi_r\, w_r \delta_r - \sum_{r \in \mathcal{R}} \pi_r\, (q_r - w_r) \tag{2}$$

Constraints (3)-(6) define a feasible cyclic route on the sequence-space network. Constraints (7)-(9) separate service execution from node visitation. The binary variable $s_r$ indicates whether request $r$ is served, while $v^P_{r,t}$ and $v^D_{r,t}$ indicate whether its pickup and delivery services are executed at position $t$. Thus, a served request must execute one pickup and one delivery action, and these actions can occur only when the corresponding logical nodes are selected. The bound on $w_r$ allows optional and partial fulfillment.

$$\text{s.t.} \quad \sum_{i\in\mathcal{V}} y_{i,t} = 1, \qquad \forall t \in \mathcal{T} \tag{3}$$

$$\sum_{j\in\mathcal{V}} x_{ij,t} = y_{i,t} \qquad \forall i \in \mathcal{V}, \forall t \in \mathcal{T} \tag{4}$$

$$\sum_{i\in\mathcal{V}} x_{ij,t} = y_{j,t+1}, \quad \sum_{i\in\mathcal{V}} x_{ij,\bar{T}} = y_{j,1} \qquad \forall j \in \mathcal{V}, \forall t \in \mathcal{T}\setminus\{\bar{T}\} \tag{5}$$

$$x_{ij,t} = 0 \qquad \forall (i,j) \in \mathcal{I}, \forall t \in \mathcal{T} \tag{6}$$

$$\sum_{t\in\mathcal{T}} v^P_{r,t} = s_r, \quad \sum_{t\in\mathcal{T}} v^D_{r,t} = s_r \qquad \forall r \in \mathcal{R} \tag{7}$$

$$v^P_{r,t} \le y_{p_r,t}, \quad v^D_{r,t} \le y_{d_r,t} \qquad \forall r \in \mathcal{R}, \forall t \in \mathcal{T} \tag{8}$$

$$0 \le w_r \le q_r s_r \qquad \forall r \in \mathcal{R} \tag{9}$$

Constraints (10)-(11) define the time structure induced by the selected sequence. As illustrated in Fig. 1, the columns in the lower matrix correspond to the service positions used in the upper time diagram. These columns are not fixed time points; instead, the real elapsed time $B_t$ at each selected logical node is accumulated along the chosen physical movements. $C$ denotes the total cycle length. Constraints (12)-(14) then assign service timestamps $T^P_r$ and $T^D_r$ to the selected pickup and delivery actions. If request $r$ is not served, these timestamps are forced to zero.

$$B_1 = 0, \quad B_{t+1} = B_t + \sum_{i\in\mathcal{V}}\sum_{j\in\mathcal{V}} \tau_{\phi(i)\phi(j)} x_{ij,t} \qquad \forall t \in \mathcal{T}\setminus\{\bar{T}\} \tag{10}$$

$$C = B_{\bar{T}} + \sum_{i\in\mathcal{V}}\sum_{j\in\mathcal{V}} \tau_{\phi(i)\phi(j)} x_{ij,\bar{T}}, \quad 0 \le C \le C_{max} \tag{11}$$

$$T^P_r \le B_t + M(1 - v^P_{r,t}), \quad T^P_r \ge B_t - M(1 - v^P_{r,t}) \qquad \forall r \in \mathcal{R}, \forall t \in \mathcal{T} \tag{12}$$

$$T^D_r \le B_t + M(1 - v^D_{r,t}), \quad T^D_r \ge B_t - M(1 - v^D_{r,t}) \qquad \forall r \in \mathcal{R}, \forall t \in \mathcal{T} \tag{13}$$

$$0 \le T^P_r \le M s_r, \quad 0 \le T^D_r \le M s_r \qquad \forall r \in \mathcal{R} \tag{14}$$

Constraints (15)-(20) define service positions, cyclic order, and soft due-date commitment tardiness. The variables $P^P_r$ and $P^D_r$ record the pickup and delivery positions of request $r$. Because the route is cyclic, a delivery may appear before its paired pickup within the reference cycle; the binary variable $z_r$ distinguishes the two cases, with $z_r = 1$ for within-cycle (pickup-before-delivery) service and $z_r = 0$ for cross-cycle (delivery-before-pickup) service. The offset $C_r$ in (18) adds one cycle length in the cross-cycle case, so that the elapsed service time $u_r$ in (19) is measured consistently. As shown in the upper panel of Fig. 1, the due-date commitment $H_r$ is a relative service horizon starting from the pickup time. Therefore, tardiness $\delta_r$ in (20) is determined by the excess of $u_r$ over $H_r$, rather than by the absolute arrival time at the delivery node.

$$P_r^P = \sum_{t\in\mathcal{T}} t\, v_{r,t}^P, \quad P_r^D = \sum_{t\in\mathcal{T}} t\, v_{r,t}^D \qquad \forall r\in\mathcal{R} \tag{15}$$

$$P_r^D - P_r^P \geq 1 - M(1-z_r) - M(1-s_r) \qquad \forall r\in\mathcal{R} \tag{16}$$

$$P_r^P - P_r^D \geq 1 - Mz_r - M(1-s_r) \qquad \forall r\in\mathcal{R} \tag{17}$$

$$C_r \geq C - Mz_r, \quad C_r \leq M(1-z_r), \quad C_r \leq C, \quad C_r \geq 0 \qquad \forall r\in\mathcal{R} \tag{18}$$

$$u_r = T_r^D - T_r^P + C_r \qquad \forall r\in\mathcal{R} \tag{19}$$

$$\delta_r \geq u_r - H_r - M(1-s_r), \quad 0 \leq \delta_r \leq C_{max} \qquad \forall r\in\mathcal{R} \tag{20}$$

Capacity feasibility is enforced by tracking cross-cycle onboard cargo and load evolution. Constraint (21) defines $h_r$, the quantity already onboard at the beginning of the reference cycle. Constraints (22)-(23) link the fulfilled quantity $w_r$ to the executed actions through $a_{r,t}^P$ and $a_{r,t}^D$. Constraints (24)-(25) initialize and propagate the onboard load, and hold the capacity limit $Q$.

$$0 \leq h_r \leq w_r, \quad h_r \leq q_r(1-z_r), \quad h_r \geq w_r - q_r z_r \qquad \forall r\in\mathcal{R} \tag{21}$$

$$0 \leq a_{r,t}^P \leq w_r, \quad a_{r,t}^P \leq q_r v_{r,t}^P, \quad a_{r,t}^P \geq w_r - q_r(1-v_{r,t}^P) \qquad \forall r\in\mathcal{R}, \forall t\in\mathcal{T} \tag{22}$$

$$0 \leq a_{r,t}^D \leq w_r, \quad a_{r,t}^D \leq q_r v_{r,t}^D, \quad a_{r,t}^D \geq w_r - q_r(1-v_{r,t}^D) \qquad \forall r\in\mathcal{R}, \forall t\in\mathcal{T} \tag{23}$$

$$L_1 = \sum_{r\in\mathcal{R}} h_r, \quad L_{t+1} = L_t + \sum_{r\in\mathcal{R}} a_{r,t}^P - \sum_{r\in\mathcal{R}} a_{r,t}^D \qquad \forall t\in\mathcal{T}\setminus\{\bar{T}\} \tag{24}$$

$$0 \leq L_t \leq Q \qquad \forall t\in\mathcal{T} \tag{25}$$

Finally, constraints (26)-(27) specify the variable domains, where all variables are indexed over their natural domains.

$$y_{i,t}, x_{ij,t}, v_{r,t}^P, v_{r,t}^D, s_r, z_r \in \{0,1\} \qquad \forall i,j\in\mathcal{V},\ r\in\mathcal{R},\ t\in\mathcal{T} \tag{26}$$

$$w_r, h_r, a_{r,t}^P, a_{r,t}^D, B_t, L_t, T_r^P, T_r^D, P_r^P, P_r^D, C, C_r, u_r, \delta_r \geq 0 \qquad \forall r\in\mathcal{R},\ t\in\mathcal{T} \tag{27}$$

However, the mixed-integer nonlinear programming (MINLP) model remains computationally challenging. Even when solving the McCormick-relaxed MIP to obtain an upper bound, the solver must branch over complex route sequences while simultaneously determining fulfilled quantities, dynamic load evolution, and soft due-date commitments. In particular, optional fulfillment and cross-cycle service make the solution space even larger, more complex, and harder to decompose. We therefore recast the problem as a finite-horizon constrained Markov decision process (CMDP), in which each action selects the next logical node and the simulator updates the operational state accordingly, providing a sequential decision-making formulation for the learning-based solution method. The full CMDP formulation is provided in Appendix A in the Online Supplement.

## 3. Framework

This section presents DCGA, a simulator-coupled reinforcement learning framework for optimizing the objective in Eq. (2). One of the central difficulties is that routing decisions and cargo flow allocation are tightly coupled under operational constraints. We therefore implement cargo flow allocation as a deterministic, rule-based simulator that guarantees operational feasibility and provides transparent state transitions. For a decoded route, the simulator evaluates a feasible cargo flow allocation, so the policy learns route sequences according to their allocation outcomes.

### 3.1. Overview

Fig. 2 summarizes the overall architecture. DCGA encodes each instance through two complementary graph channels: the network channel captures sparse reachability and non-Euclidean arc attributes, while the demand channel captures OD coupling and service information. This separation prevents physical-movement features and demand-service logic from being mixed prematurely. The two representations are then fused into static node embeddings, which provide a unified feature for the state-conditioned decoder.

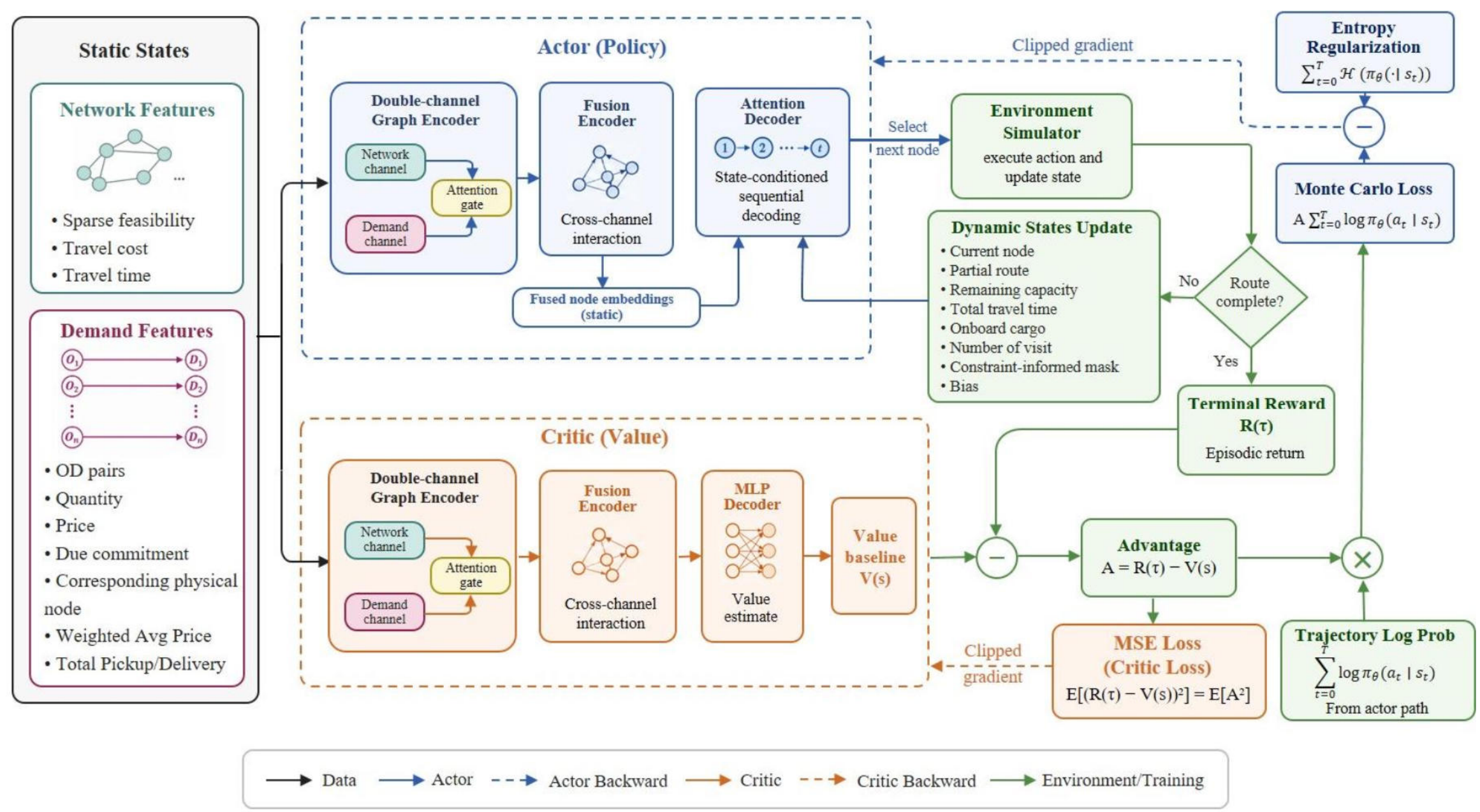

**Figure 2** **Overall architecture of the proposed DCGA framework.**

Because route desirability changes with location, time, capacity, onboard cargo, and service status, the decoder constructs the service route autoregressively while the environment updates the operational state after each selected action. After route completion, the simulator evaluates the terminal objective through feasible cargo flow allocation, so the actor is trained according to downstream profit rather than travel cost alone. A critic provides a variance-reduction baseline for actor-critic training.

## 3.2. Actor Framework

We now detail the actor, which implements the autoregressive route-construction policy. At each decoding step $t$, the actor selects the next service action according to $\pi_\theta(a_t \mid s_t)$, where $s_t$ is the current state and $a_t$ is the selected logical node. The actor consists of a double-channel encoder, a fusion encoder, a state-conditioned decoder, and a constraint-informed policy-shaping module.

### 3.2.1. Representation design for sparse non-Euclidean routing

A key challenge in the NE-PDP-CCS is that routing feasibility cannot be inferred from planar coordinates: the network is sparse, directed, asymmetric, and non-Euclidean. Moreover, each action carries not only movement information but also OD-specific service role, quantity, reward, and due-date commitment attributes. We therefore use the logical node representation defined in Sec. 2 as the action space, which also allows the decoder to track visited services and paired pickup-delivery status for masks and delivery-side bias. For policy construction, we add a zero-cost virtual node 0 to both $\mathcal{V}$ and $\mathcal{K}$, and use the same notation for the augmented sets. This node initializes route construction and lets the actor choose the first physical service location endogenously.

Motivated by these structural features, we design a double-channel graph encoder, as shown in Fig. 3. The network channel captures feasible physical movements and arc-level travel attributes, whereas the demand channel captures OD coupling and service-role information. Their outputs are then fused into a static node tensor used by the sequential decoder.

Specifically, we represent each instance using two complementary graph channels. The first channel is the network graph $G_v = (\mathcal{V}_v, E_v)$, which encodes physical reachability and arc-level operational attributes. Its arc attributes include travel cost, travel time, and feasibility information. The second channel is the demand graph $G_p = (\mathcal{V}_p, E_p)$, which encodes OD coupling and service information. Arcs in this graph connect pickup and delivery logical nodes and carry demand-side attributes such as quantity, reward, and soft due-date commitment information.

However, logical nodes alone are not sufficient for the demand channel, because multiple OD-specific service roles may map to the same physical node. For the network channel, this does not require special treatment: reachability, travel cost, and travel time are determined by physical-node pairs and can be inherited by each mapped logical node. For the demand channel, however, encoding only the original OD-pair arcs may fragment the demand-service relations associated with a multi-task physical node. We therefore introduce logical-physical-logical (LPL) projection, which aggregates demand-side logical features to the physical level, encodes the aggregated representation on the physical graph, and maps the learned embeddings back to the logical action space. An illustrative example is provided in Appendix B in the Online Supplement. Formally, let $\phi : \mathcal{V} \to \mathcal{K}$ denote the logical-to-physical mapping, and let $\Pi \in \{0,1\}^{|\mathcal{V}|\times|\mathcal{K}|}$ be the corresponding logical-physical incidence matrix:

$$\Pi_{i,k} = \mathbb{I}[\phi(i) = k], \qquad n = \Pi^\top \mathbf{1}_{|\mathcal{V}|}, \qquad \Lambda = \mathrm{diag}(n), \tag{28}$$

where $n$ counts how many logical nodes are mapped to each physical node and $\Lambda$ is the normalization matrix. Given the demand node-feature tensor $X_p$, logical-physical projection (LPP) aggregates logical features into physical-level features:

$$\bar{X}_p = \mathrm{LPP}(X_p) := \Lambda^{-1}\Pi^\top X_p. \tag{29}$$

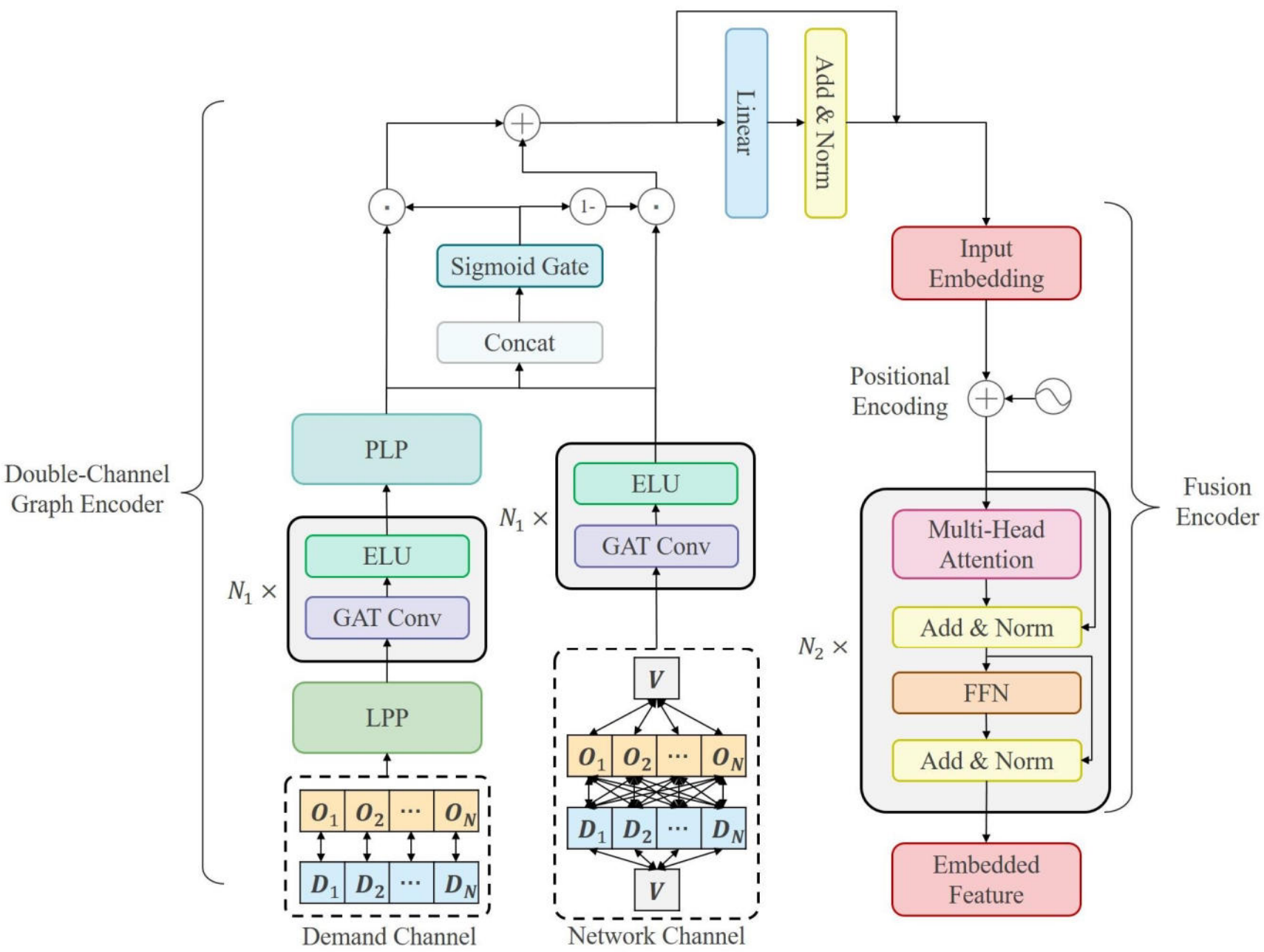

**Figure 3** **Architecture of the double-channel graph encoder.**

The aggregated representation is then encoded on the physical graph and mapped back to the logical action space through physical-logical projection (PLP):

$$\bar{H}_p = \mathcal{G}_p(\bar{X}_p, \bar{E}_p), \qquad H_p = \mathrm{PLP}(\bar{H}_p) := \Pi\bar{H}_p, \tag{30}$$

where $\mathcal{G}_p$ denotes the $N_1$-layer physical GAT encoder with arc features $\bar{E}_p$. Together, LPP, physical GAT encoding, and PLP form the LPL projection, enabling information sharing among logical service roles located at the same physical node while preserving the OD-indexed action space used by the decoder.

The network channel is encoded directly in the logical action space. Because reachability, travel cost, and travel time are inherited from the mapped physical nodes, no logical-physical aggregation is required for this channel. Starting from the network node-feature tensor $X_v$, we apply $N_1$ GAT-ELU layers on the network graph with arc features $E_v$:

$$H_v^{(0)} = X_v, \qquad H_v^{(l+1)} = \mathrm{ELU}\big(\mathrm{GAT}(H_v^{(l)}, E_v)\big), \quad l = 0, \dots, N_1 - 1, \qquad H_v := H_v^{(N_1)}. \tag{31}$$

#### 3.2.2. Global fusion for inter-channel interaction

Together with the demand-channel embedding $H_p$, these two views must be fused before decoding, because each routing decision simultaneously determines the next physical movement and the downstream cargo flow consequences. We therefore use a learnable gate to combine the two channel embeddings and incorporate the combined node-feature tensor $X$ (includes $X_p$ and $X_v$) together:

$$g = \sigma\big([H_v; H_p] W_g + b_g\big), H_g = g \odot H_v + (1 - g) \odot H_p, \tag{32}$$

$$H^{(0)} = \mathrm{LN}\big(H_g W_0 + X W_{\mathrm{res}}\big). \tag{33}$$

The gated representation integrates the two local graph views, but route construction also requires global comparison across distant nodes and OD pairs. We therefore refine the fused node tensor using $N_2$ Transformer-style attention layers:

$$\hat{H}^{(\ell)} = \mathrm{LN}\left(H^{(\ell-1)} + \mathrm{MHA}\big(H^{(\ell-1)}, H^{(\ell-1)}, H^{(\ell-1)}\big)\right), \tag{34}$$

$$H^{(\ell)} = \mathrm{LN}\left(\hat{H}^{(\ell)} + \mathrm{FFN}(\hat{H}^{(\ell)})\right), \qquad \ell = 1, \dots, N_2. \tag{35}$$

The final static node tensor and graph-level summary are then computed as

$$\bar{h} = \frac{1}{|\mathcal{V}|} \sum_{i \in \mathcal{V}} H_{:,i,:}^{(N_2)}, \tag{36}$$

and $H^{(N_2)}$ and $\bar{h}$ are passed to the decoder.

#### 3.2.3. State-conditioned sequential decoding

During decoding, the value of a candidate node depends on the evolving operational state, including the current location, residual capacity, remaining time, onboard cargo, and residual demand. We therefore condition each decision on both the static node tensor $H^{(N_2)}$ produced by the encoder and the dynamic state $s_t = (c_t, p_t, x_t^{\mathrm{cap}}, x_t^{\mathrm{time}})$, where $c_t$ is the current logical node, $p_t = (a_0, \dots, a_{t-1})$ is the partial route, and $x_t^{\text{cap}}$ and $x_t^{\text{time}}$ are the residual capacity and remaining time provided by the simulator.

As shown in Fig. 4, the decoder uses two lightweight state-conditioning modules. The micro-encoder summarizes the ordered partial route into a history representation, while the conditional attention gate modulates candidate-node embeddings according to the current time-capacity state.

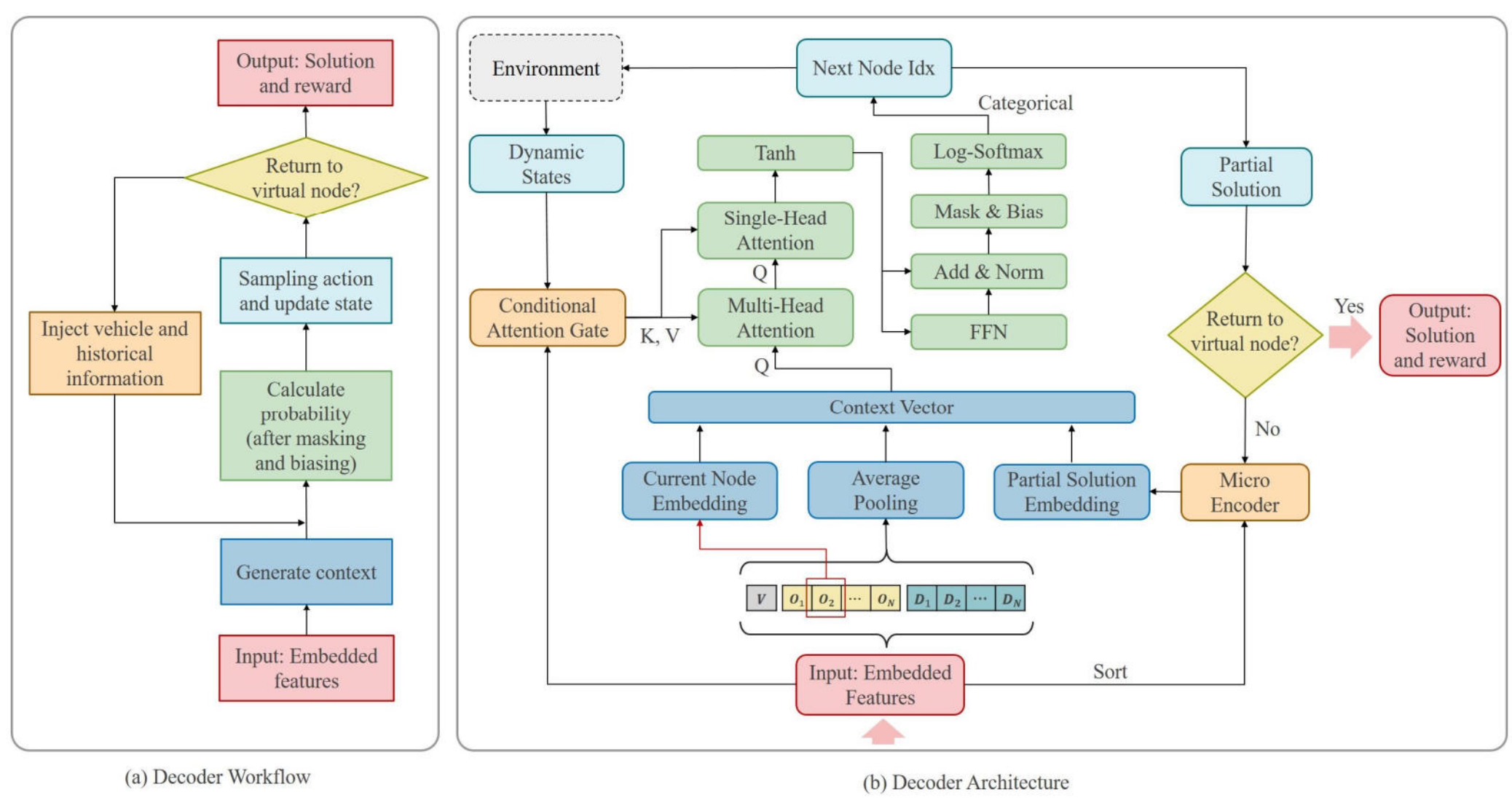

**Figure 4** **Actor decoder flowchart and architecture.**

Together with the graph-level summary $\bar{h}$ and the current-node embedding, these components produce a state-conditioned context $h_t^c$; details are given in Appendix D in the Online Supplement. Given $h_t^c$, the raw score of candidate node $j$ is $U_t$:

$$U_{t,j} = w_u^\top \tanh\left(\gamma\,[h_t^c \| H^{(N_2)}[:, j, :]]\right) + b_u, \qquad j \in \mathcal{V}, \tag{37}$$

where $w_u$, $b_u$, and $\gamma$ are trainable scoring parameters.

#### 3.2.4. Constraint-informed policy shaping

On sparse non-Euclidean networks, unconstrained exploration may lead to disconnected moves, capacity-infeasible pickups, excessive revisits, or invalid stopping actions. We therefore shape the raw decoder scores $U_t$ by combining a finite delivery-side bias with hard operational masks:

$$\tilde{U}_t = U_t + \xi_t + \mathcal{M}_t, \qquad \pi_\theta(a_t = j \mid s_t) = \text{softmax}(\tilde{U}_t)_j, \quad j \in \mathcal{V}, \tag{38}$$

where $\xi_t \in \mathbb{R}^{B\times|\mathcal{V}|}$ is a finite bias term, $B$ is the batch size, and $\mathcal{M}_t \in \{-\infty, 0\}^{B\times|\mathcal{V}|}$ is the hard operational mask. The finite bias discourages delivery-type actions whose paired pickup has not yet been visited, while still allowing such actions under cyclic cross-cycle interpretation. Let $S_t^{(b)}$ denote the set of logical nodes visited by batch element $b$, and let $p(j) = j - N$ be the paired pickup node of delivery node $j \in \mathcal{V}^D$. Denoting $\eta_D$ as the bias coefficient, we define

$$(\xi_t)_{b,j} = -\eta_D \, \mathbb{I}[j \in \mathcal{V}^D] \, \mathbb{I}[p(j) \notin S_t^{(b)}], \qquad \eta_D \geq 0. \tag{39}$$

The hard mask is decomposed into four components:

$$\mathcal{M}_t = \mathcal{M}_t^{\text{leg}} + \mathcal{M}_t^{\text{cap}} + \mathcal{M}_t^{\text{visit}} + \mathcal{M}_t^{\text{stop}}. \tag{40}$$

Each component takes value 0 when the corresponding condition is satisfied and $-\infty$ otherwise. Specifically, the leg mask enforces sparse reachability, the capacity mask rules out capacity-infeasible pickups, the visit mask limits repeated visits to the same physical node, and the stop mask allows selecting the virtual node only when termination is admissible:

$$\mathcal{M}_{t,b,j}^{\text{leg}} = 0 \iff (c_{t,b}, j) \in E_v, \tag{41}$$

$$\mathcal{M}_{t,b,j}^{\text{cap}} = 0 \iff j \notin \mathcal{V}^P \text{ or } x_{t,b}^{\text{cap}} + \Delta_{t,b}^{\text{unload}}(j) \geq q_{t,b}^{\min}(j), \tag{42}$$

$$\mathcal{M}_{t,b,j}^{\text{visit}} = 0 \iff j = 0 \text{ or } n_{t,b}(\phi(j)) < \bar{n}, \tag{43}$$

$$\mathcal{M}_{t,b,j}^{\text{stop}} = 0 \iff j \neq 0 \text{ or } \text{CanReturn}(s_{t,b}) = 1, \tag{44}$$

Here, $c_{t,b}$ is the current logical node, $x_{t,b}^{\text{cap}}$ is the residual capacity, $\Delta_{t,b}^{\text{unload}}(j)$ is the quantity that can be unloaded before serving candidate node $j$, $q_{t,b}^{\min}(j)$ is the minimum quantity required for a feasible pickup, $n_{t,b}(\phi(j))$ is the number of previous visits to physical node $\phi(j)$, $\bar{n}$ is the visit-limit threshold, and the indicator $\text{CanReturn}(s_{t,b})$ equals one when either the number of remaining unvisited physical service nodes is no larger than a threshold $\bar{m}$, or no feasible non-virtual action remains. This condition only permits the actor to select the virtual node 0; the route is closed if this return action is selected. Together, the finite bias and hard masks preserve cyclic-service flexibility while removing structurally invalid actions from the policy support.

### 3.3. Critic Framework

The critic is used as a variance-reduction baseline in episodic actor-critic training. It adopts a single-layer version of the actor encoder to obtain a graph-level representation and maps this representation to an instance-level value estimate $V_\psi(H^{(N_2)'})$ through a three-layer MLP. The critic is trained to predict the Monte Carlo terminal return and does not affect the decoding policy at inference time. Detailed architecture and estimation equations are provided in Appendix E in the Online Supplement.

### 3.4. Training Algorithm

Non-Euclidean PDP (NE-PDP) has two properties that make step-wise reward learning difficult. First, feasibility and profit are determined by the entire decoded route rather than by isolated actions.

Second, sparse connectivity can delay the consequences of early routing decisions, making bootstrapped value targets noisy. We therefore train DCGA with an episodic objective: the route is evaluated only after completion, the resulting Monte Carlo return is used as the policy-gradient signal, and a critic baseline is introduced to reduce variance.

For each instance $i$, the actor samples a trajectory $\tau_i = (a_0^{(i)}, \ldots, a_{\bar{T}_i-1}^{(i)})$ from the masked policy. After termination, the simulator evaluates the terminal objective $R_i = R(\tau_i)$ according to Eq. (2). The critic provides a state-dependent baseline $V_\psi(H^{(N_2)'})$, and the advantage is computed as

$$A_i = R_i - V_\psi(H^{(N_2)'}). \tag{45}$$

The agent is optimized by

$$\mathcal{L}_\pi(\theta) = -\frac{\alpha}{B}\sum_{i=1}^{B}\sum_{t=0}^{\bar{T}-1} A_i \log \pi_\theta(a_t^{(i)} \mid s_t^{(i)}) - \beta \frac{1}{B}\sum_{i=1}^{B}\sum_{t=0}^{\bar{T}-1} \mathcal{H}\Big(\pi_\theta(\cdot \mid s_t^{(i)})\Big), \tag{46}$$

$$\mathcal{L}_v(\psi) = \frac{1}{2B}\sum_{i=1}^{B}\big(V_\psi(H^{(N_2)'}) - R_i\big)^2. \tag{47}$$

The entropy term in Eq. (46) encourages exploration over the masked feasible action set. Appendix C of the Online Supplement gives a variational interpretation. At inference time, when using the pretrained model, we replace sampling with greedy selection. In addition, to stabilize optimization, we clip the $\ell_2$-norm of the gradient during backpropagation. Gradient clipping bounds the overall update magnitude and reduces the risk of unstable parameter updates.

## 4. Numerical Experiments

We design the experiments to evaluate DCGA from five perspectives: solution quality against strong benchmarks on small and large batches (Secs. 4.2-4.3), computational efficiency on large batches where iterative methods become expensive (Sec. 4.3), the effectiveness of key architectural mechanisms such as the double-channel representation and feasibility-aware masks (Sec. 4.4), stability and generalization under demand perturbations and instance variations (Sec. 4.5), and sensitivity to key hyperparameters (Sec. 4.6 and Appendix I in the Online Supplement).

We use liner-shipping network design as a representative testbed because, as discussed in Sec. 1, it naturally exhibits the key structural features of the NE-PDP-CCS. The instances are constructed from the Mediterranean subset of LinerLib (Brouer et al. 2014) (available at `https://github.com/blof/LINERLIB`), a public benchmark suite based on realistic large-scale liner-shipping network data. This testbed preserves realistic network topology and operational attributes while allowing controlled generation of OD requests and demand parameters across instance scales. Details on instance construction, demand generation, model hyperparameters, and benchmark settings are provided in Appendix F in the Online Supplement.

### 4.1. Benchmarks

To evaluate the proposed DCGA framework, we compare it with three classes of benchmarks: exact and decoupled optimization models, classical heuristic and metaheuristic methods, and adapted learning-based methods. Detailed implementation hyperparameters are reported in Appendix F in the Online Supplement. Table 2 summarizes the construction and role of each benchmark.

**Table 2 Overview of benchmark methods.**

| Category | Method | Main construction | Role |
|---|---|---|---|
| Optimization | MIP | Uses Gurobi to solve the McCormick-relaxed MIP derived from the exact formulation in Sec. 2, with the relaxation detailed in Appendix G. Given sufficient time, it could provide an optimistic upper-bound optimization benchmark. | Relaxed-MILP benchmark. |
| | Flow model (FM) | First solves a TSP-style MIP to obtain a minimum-cost feasible service loop; then fixes the sequence and solves a linear program for cargo flow allocation. | Decoupled optimization benchmark. |
| Heuristic and metaheuristic | LKH3 (Helsgaun 2017) | Uses the closest supported LKH3 variant, m-PDTSP, as the routing core. Additional adaptations are summarized in Table S.5. | Strong classical routing benchmark. |
| | $GA_1$ (Jung and Haghani 2000) | Uses genetic search with travel-cost minimization as the routing objective. | Efficiency-oriented population search. |
| | $GA_2$ (Jung and Haghani 2000) | Same as $GA_1$, but uses the full reward function. | Quality-oriented population search. |
| | $SA_1$ (Li and Lim 2001) | Uses 2-opt neighborhoods with simulated-annealing acceptance and travel-cost minimization as the routing objective. | Efficiency-oriented local search. |
| | $SA_2$ (Li and Lim 2001) | Same as $SA_1$, but uses the full reward function as the search objective. | Quality-oriented local search. |
| | $ALNS_1$ (Ropke and Pisinger 2006) | Uses relocate, swap, segment reversal, insertion, and deletion operators with adaptive operator weights, initialized from a random construction. | Adaptive neighborhood search. |
| | $ALNS_2$ (Ropke and Pisinger 2006) | Same as $ALNS_1$, but starts from a value-greedy construction. | Adaptive neighborhood search. |
| Learning-based | $NCS_1$ (Kong et al. 2024) | Adapts NCS, which strengthens N2S with an RL- and imitation-learning-based initial solution constructor, by using port longitude and latitude as coordinate inputs and adding a feasibility mask for sparse-network connectivity. | Strong coordinate-based neural PDP reference. |
| | $NCS_2$ (Kong et al. 2024) | Same as $NCS_1$, but adds the same graph encoder used in DCGA before the NCS backbone while retaining the original NCS decoding framework. Additional construction details are summarized in Table S.6. | Graph-enhanced neural PDP reference. |

Kong et al. (2024) showed that NCS achieves leading performance among learning-based methods on PDP benchmarks. Since NCS shares the Euclidean-coordinate representation assumptions commonly adopted by other neural PDP solvers, we adapt its variants as representative learning-based benchmarks and do not additionally include MAAM (Zhang et al. 2020), Joint T-RL (Zhang et al. 2022), or HAM (Santiyuda et al. 2024).

### 4.2. Analysis on Small-Batch Instances

To validate the effectiveness of the proposed DCGA framework, we conduct a comprehensive comparative analysis against the benchmarks. Because solving the full MIP to global optimality is prohibitively time-consuming, we impose a one-hour-per-instance time limit and report results on a small batch of 15 instances. Fig. 5 summarizes the small-batch results across different instance scales, using the relaxed model defined in Appendix G in the Online Supplement.

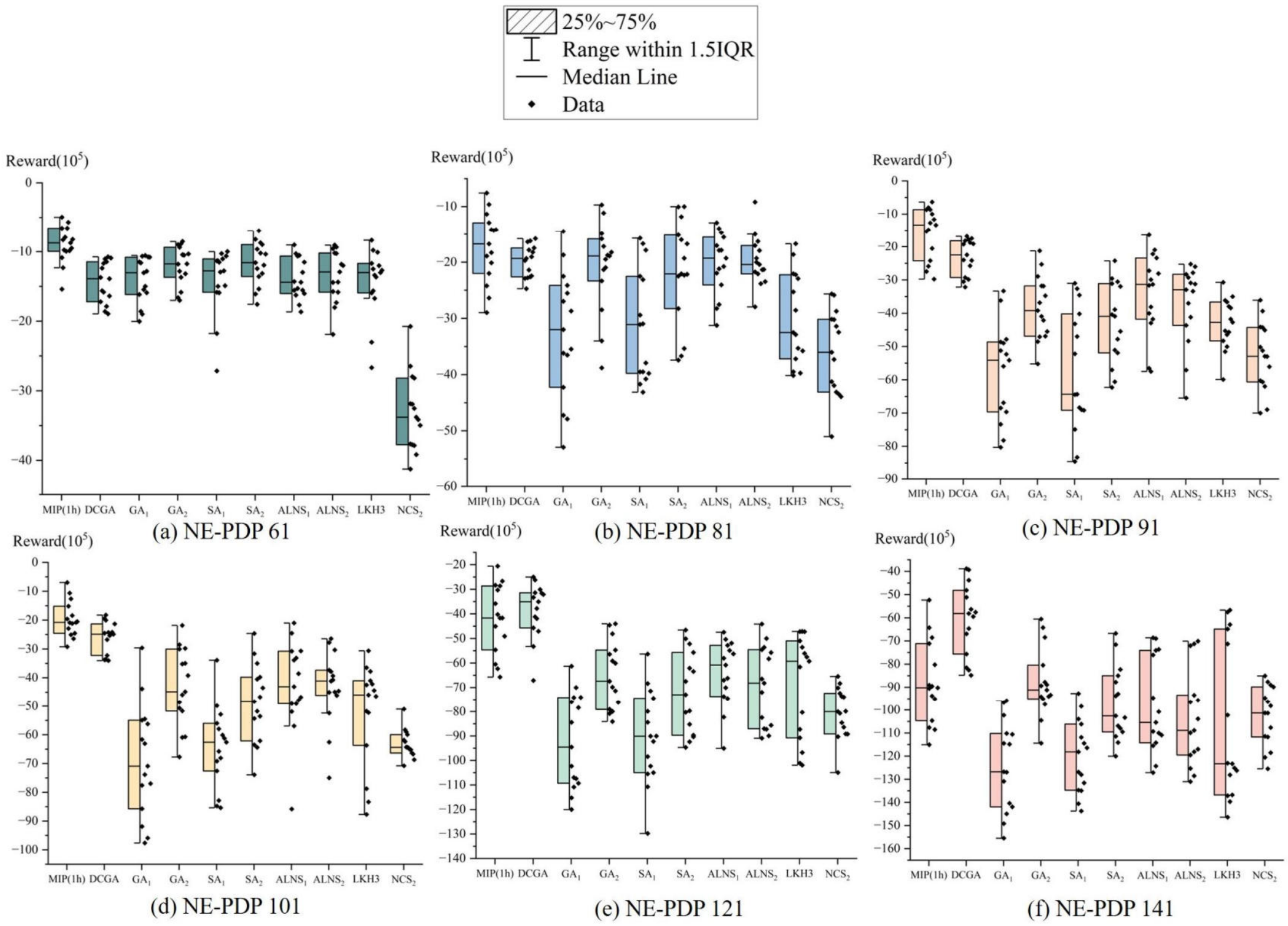

**Figure 5** **Small-batch performance comparison with time-limited MIP (1 h per instance) across instance scales.**

As the problem size increases, the MIP solver becomes increasingly constrained by the one-hour budget and its incumbent objective gradually falls behind the solutions produced by DCGA. This gap indicates that, for larger instances, the limiting factor is no longer modeling fidelity but the inability to explore the combinatorial space within a fixed budget while DCGA, as a tailored reinforcement learning method, can effectively address this issue.

### 4.3. Analysis on Large-Batch Instances

We next evaluate DCGA on larger test batches to examine whether the learned policy remains effective beyond the small-batch setting used for comparison with the time-limited MIP. The experiments also cover six instance scales and compare DCGA with optimization, heuristic, metaheuristic, and learning-based baselines. We report the mean objective value over each test batch, the total runtime for processing the entire batch, and relative gap to the best-known objective, gap(%) $= \frac{|\text{Obj}-\text{Obj}_{\text{best}}|}{|\text{Obj}_{\text{best}}|} \times$ 100%. The training curves in Fig. 6 show that DCGA converges stably across representative instance scales.

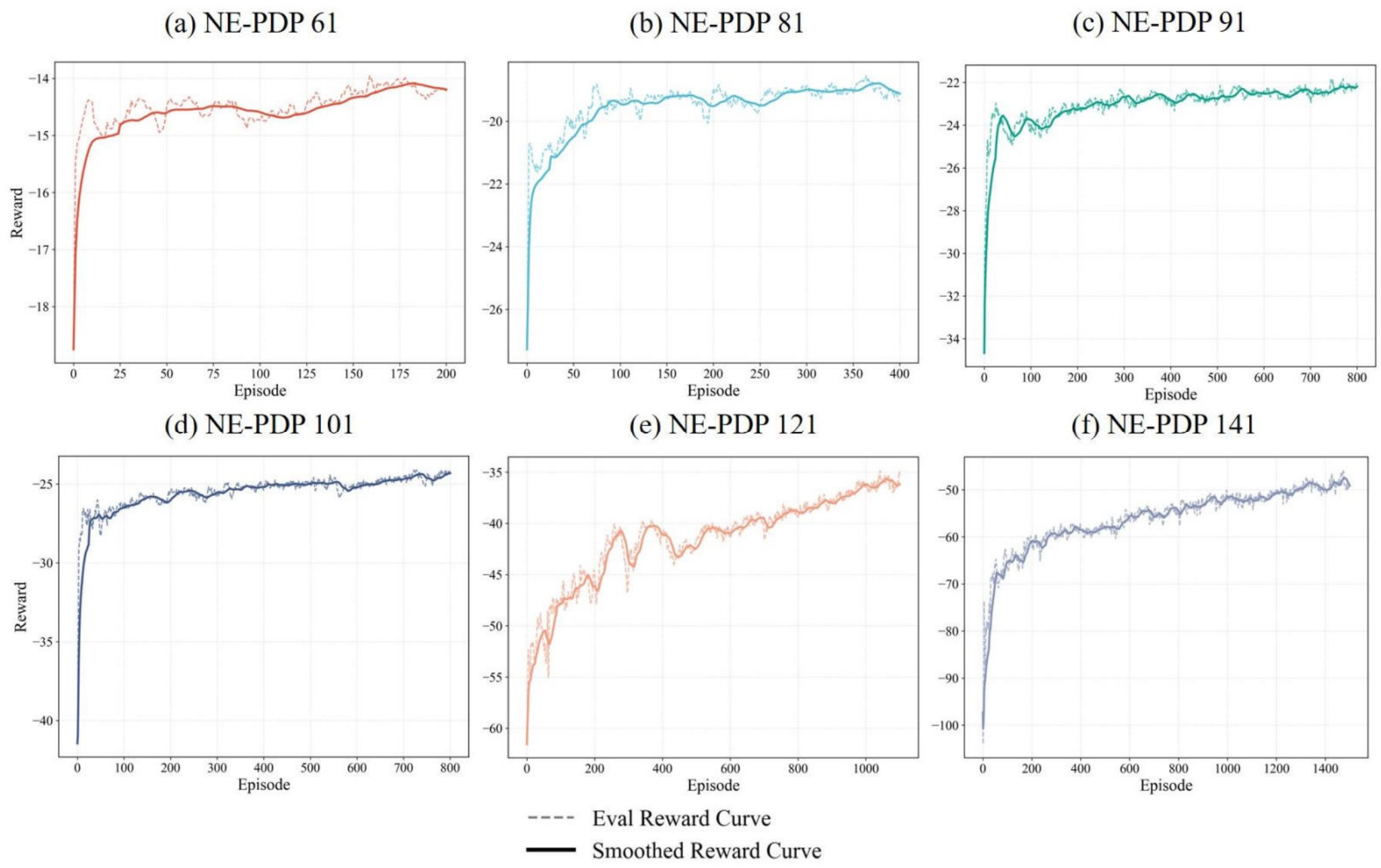

**Figure 6** Training curves across representative instance scales.

**Table 3** Performance comparison on NE-PDP 61/81/91

| Method | NE-PDP 61 Batch 600 | | | NE-PDP 81 Batch 350 | | | NE-PDP 91 Batch 350 | | |
|---|---|---|---|---|---|---|---|---|---|
| | Obj($10^5$) | Gap(%) | Runtime(s) | Obj($10^5$) | Gap(%) | Runtime(s) | Obj($10^5$) | Gap(%) | Runtime(s) |
| DCGA | -14.03 | 14.07 | **17.88** | **-18.77** | **0.00** | **15.81** | **-22.07** | **0.00** | **16.38** |
| FM | -20.87 | 69.67 | 46.67 | -39.95 | 112.84 | 43.45 | -53.81 | 143.82 | 64.30 |
| $ALNS_1$ | -13.53 | 10.00 | 195.13 | -21.74 | 15.82 | 189.59 | -30.14 | 36.57 | 283.06 |
| $ALNS_2$ | -13.61 | 10.65 | 287.04 | -22.31 | 18.86 | 236.83 | -31.46 | 42.55 | 291.95 |
| $GA_1$ | -14.49 | 17.80 | 250.12 | -32.34 | 72.30 | 162.12 | -46.06 | 108.70 | 178.46 |
| $SA_1$ | -14.43 | 17.32 | 55.71 | -31.97 | 70.32 | 38.66 | -45.01 | 103.94 | 42.63 |
| $GA_2$ | **-12.30** | **0.00** | 691.10 | -22.29 | 18.75 | 489.58 | -32.24 | 46.08 | 663.26 |
| $SA_2$ | -12.90 | 4.88 | 450.25 | -25.45 | 35.59 | 330.51 | -37.36 | 69.28 | 456.61 |
| LKH3(#trials = 1000) | -15.54 | 26.34 | 269.18 | -31.84 | 69.63 | 1.11 h | -38.84 | 75.99 | 2.00 h |
| LKH3(#trials = 5000) | -15.42 | 25.37 | 1308.45 | -29.13 | 55.19 | 7.92 h | -38.13 | 72.77 | 10.68 h |
| $NCS_1$ | -40.90 | 232.52 | 1071.33 | -50.64 | 169.79 | 967.81 | -59.10 | 167.78 | 1807.81 |
| $NCS_2$ | -34.07 | 176.99 | 2.41 h | -40.30 | 114.70 | 2.07 h | -53.59 | 142.82 | 2.12 h |

Table 3 reports the results on small-to-medium instances. On the smallest size (NE-PDP 61), DCGA achieves the fastest inference but is slightly dominated in solution quality by $GA_2$, indicating that population search can still be competitive at smaller scales when sufficient runtime is allowed. As the instance size increases to NE-PDP 81 and NE-PDP 91, DCGA achieves the best-known objective values, while maintaining the shortest inference times. In contrast, classical search methods require substantially longer runtimes, and the adapted NCS variants perform poorly despite their learning-based design, suggesting that coordinate-based neural PDP solvers have difficulty handling sparse non-Euclidean network and routing-allocation coupling.

**Table 4 Performance comparison on NE-PDP 101/121/141**

| Method | NE-PDP 101 Batch 350 | | | NE-PDP 121 Batch 350 | | | NE-PDP 141 Batch 300 | | |
|---|---|---|---|---|---|---|---|---|---|
| | Obj($10^5$) | Gap(%) | Runtime(s) | Obj($10^5$) | Gap(%) | Runtime(s) | Obj($10^5$) | Gap(%) | Runtime(s) |
| DCGA | **-25.06** | **0.00** | **16.15** | **-35.88** | **0.00** | **20.29** | **-50.74** | **0.00** | **18.31** |
| FM | -62.99 | 151.36 | 80.32 | -89.98 | 150.78 | 120.43 | -128.26 | 152.78 | 150.97 |
| $ALNS_1$ | -36.71 | 46.49 | 286.93 | -62.36 | 73.80 | 290.25 | -99.15 | 95.41 | 150.16 |
| $ALNS_2$ | -38.08 | 51.96 | 240.94 | -67.51 | 88.15 | 427.87 | -108.97 | 114.76 | 319.88 |
| $GA_1$ | -59.26 | 136.47 | 184.98 | -91.72 | 155.63 | 212.57 | -127.51 | 151.30 | 200.41 |
| $SA_1$ | -57.21 | 128.29 | 45.01 | -88.28 | 146.04 | 50.45 | -122.92 | 142.25 | 47.66 |
| $GA_2$ | -39.47 | 57.50 | 738.95 | -62.58 | 74.41 | 917.19 | -90.72 | 78.79 | 945.25 |
| $SA_2$ | -45.46 | 81.40 | 514.38 | -74.29 | 107.05 | 642.85 | -106.90 | 110.68 | 668.97 |
| LKH3(#trials = 1000) | -44.30 | 76.78 | 3.83 h | -69.46 | 93.59 | 1.07 days | -105.79 | 108.49 | 1.50 days |
| LKH3(#trials = 5000) | -44.13 | 76.10 | 23.53 h | -67.78 | 88.91 | 3.56 days | – | – | > 5 days |
| $NCS_1$ | -65.51 | 161.41 | 1948.25 | -86.60 | 141.36 | 2374.17 | -105.37 | 107.67 | 2615.63 |
| $NCS_2$ | -62.70 | 150.20 | 2.28 h | -79.55 | 121.71 | 2.47 h | -102.48 | 101.97 | 2.86 h |

Table 4 shows that the advantage of DCGA becomes more pronounced on large-scale instances. DCGA achieves the best-known objective values on all three large scales in under 21 seconds. By comparison, baselines require much longer runtimes and suffer from degraded solution quality, while DCGA maintains stable inference within seconds.

Overall, DCGA provides the fastest response across all tested large-batch settings. Its solution quality is also consistently superior except on NE-PDP 61. Besides, as the instance size increases, the advantage of DCGA becomes more pronounced. These results support the intended role of DCGA as a fast and high-quality solver for medium- and large-scale sparse non-Euclidean routing-and-flow optimization.

### 4.4. Mechanism Validation

We further examine whether DCGA's performance gain comes from the proposed structural mechanisms through ablation experiments.

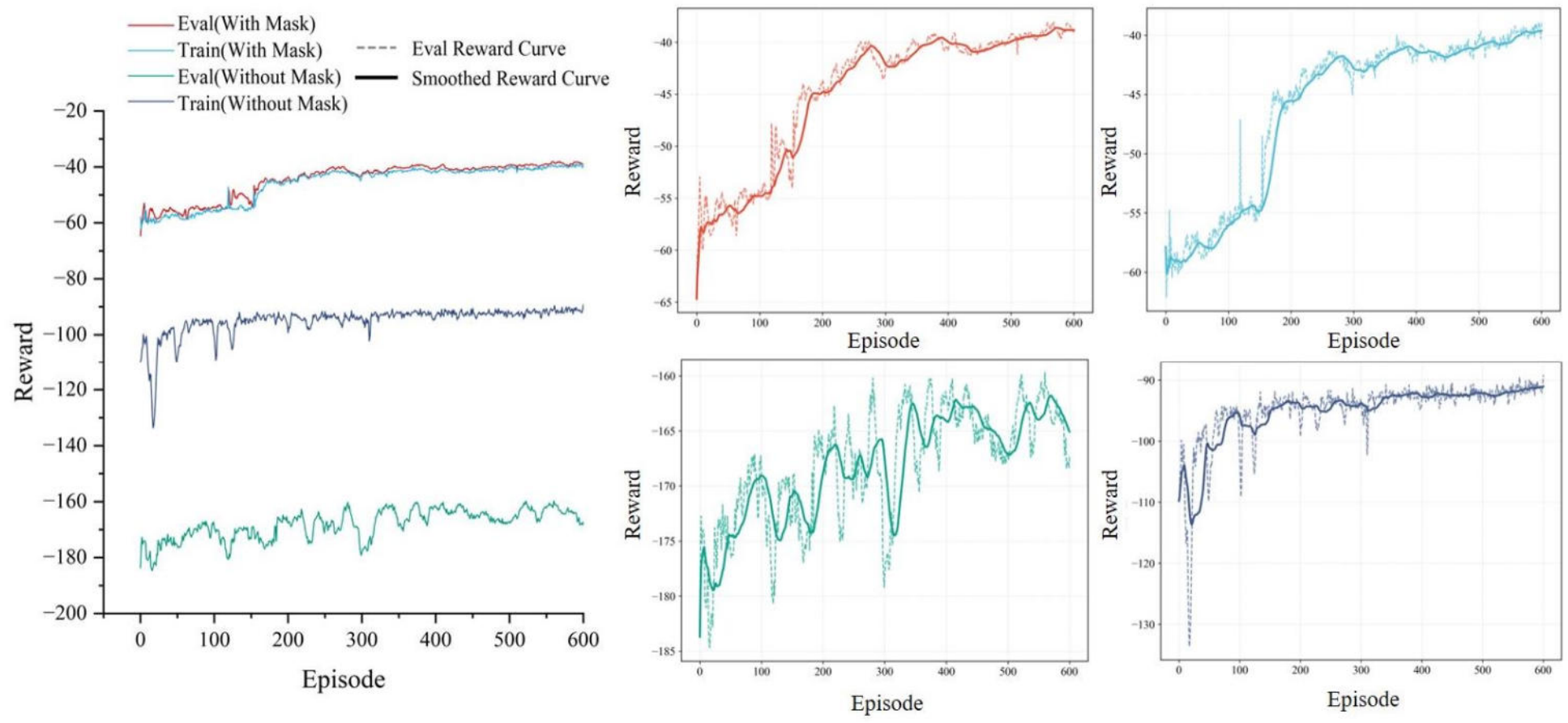

**Figure 7 Masking-mechanism ablation on NE-PDP 121.**

***Masking mechanism.*** We remove all masks *except* the basic network reachability constraint. Thus, the decoder still respects sparse-network feasibility, but no longer applies the additional masks that guide high-quality decisions during decoding. Fig. 7 shows that the full masking mechanism yields stable learning with a small train-eval gap, whereas removing the masks leads to consistently poor performance in both training and evaluation, indicating that the policy is frequently driven to low-quality decision sequences without feasibility-aware screening.

Fig. 8 further reveals that a major source of the degradation is the unfulfilled-demand penalty: without masks, unfulfilled-demand penalties increase dramatically and remain high across episodes. The ridgeline distributions also highlight that, compared with other benchmarks, a key advantage of DCGA is its strong fulfillment capability, which is largely enabled by the masking mechanism.

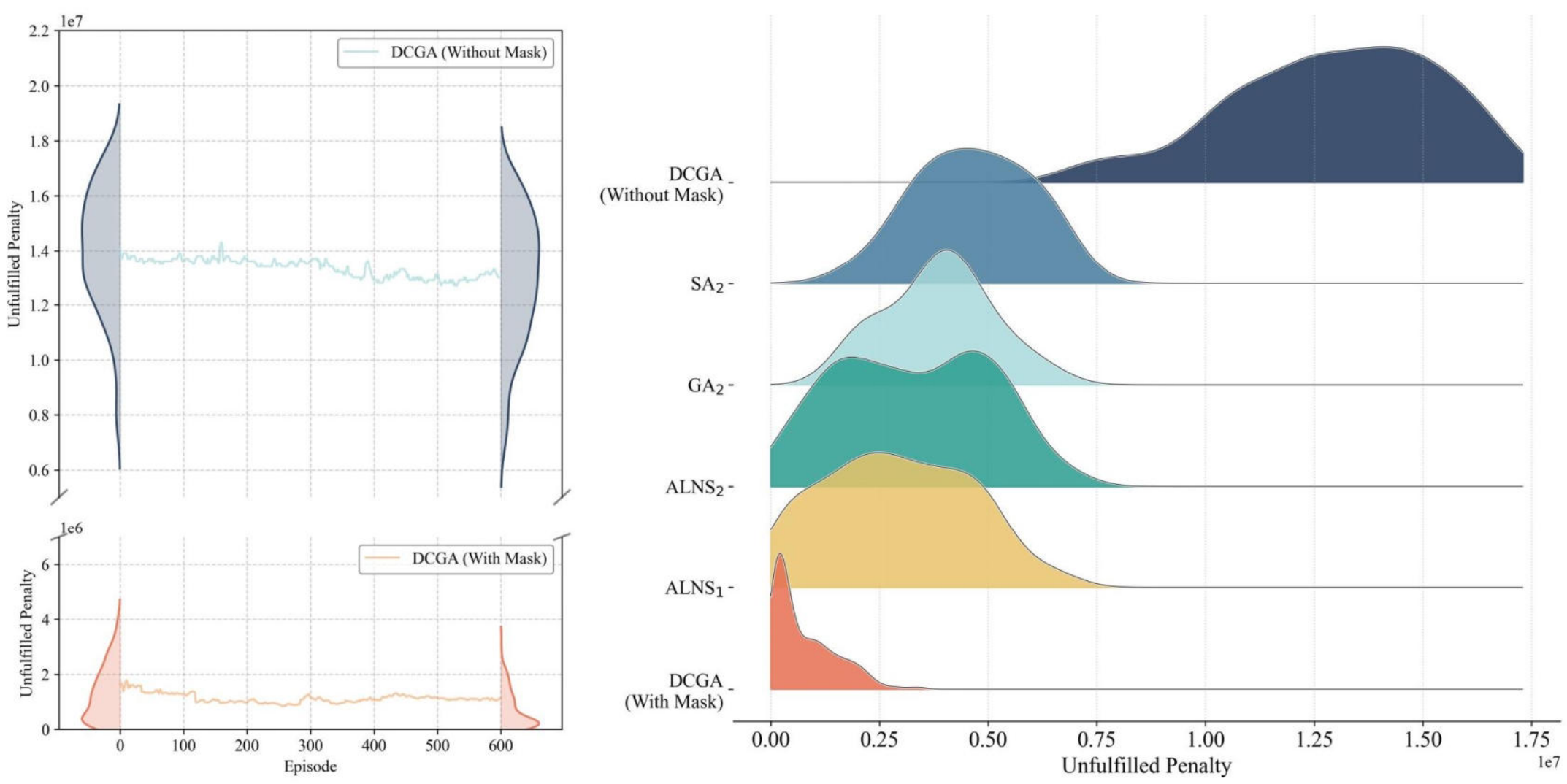

**Figure 8** **Comparative analysis of unfulfilled-demand penalties. Left: episode trends with boundary density distributions. Right: ridgeline distributions across benchmark models.**

***Number of channels.*** We further compare a *single-channel* model against our *double-channel* design under multiple scenarios. Fig. 9 presents a point-line plot summarizing the performance gap across different problem size settings. Across all tested scenarios, the DCGA model consistently dominates the Single-Channel Graph Attention (SCGA) baseline, and the advantage becomes even greater as the size of the problem increases, indicating that the additional channel provides complementary information that improves decision quality and stability.

### 4.5. Stability and Generalization Analysis

We further evaluate whether the learned policy remains effective under instance variations beyond the nominal training setting. Even with a fixed network topology, the combinatorial space induced by varying request sets and attributes is enormous. For example, in NE-PDP 91, selecting 23 task nodes from a candidate pool of 39 ports yields $\binom{39}{23} \approx 10^{10}$ possible node-subset realizations before accounting for continuous attributes such as quantities and due dates. Since training covers only a small fraction of this space, stable performance would indicate that the policy learns transferable decision rules rather than memorizing specific instances.

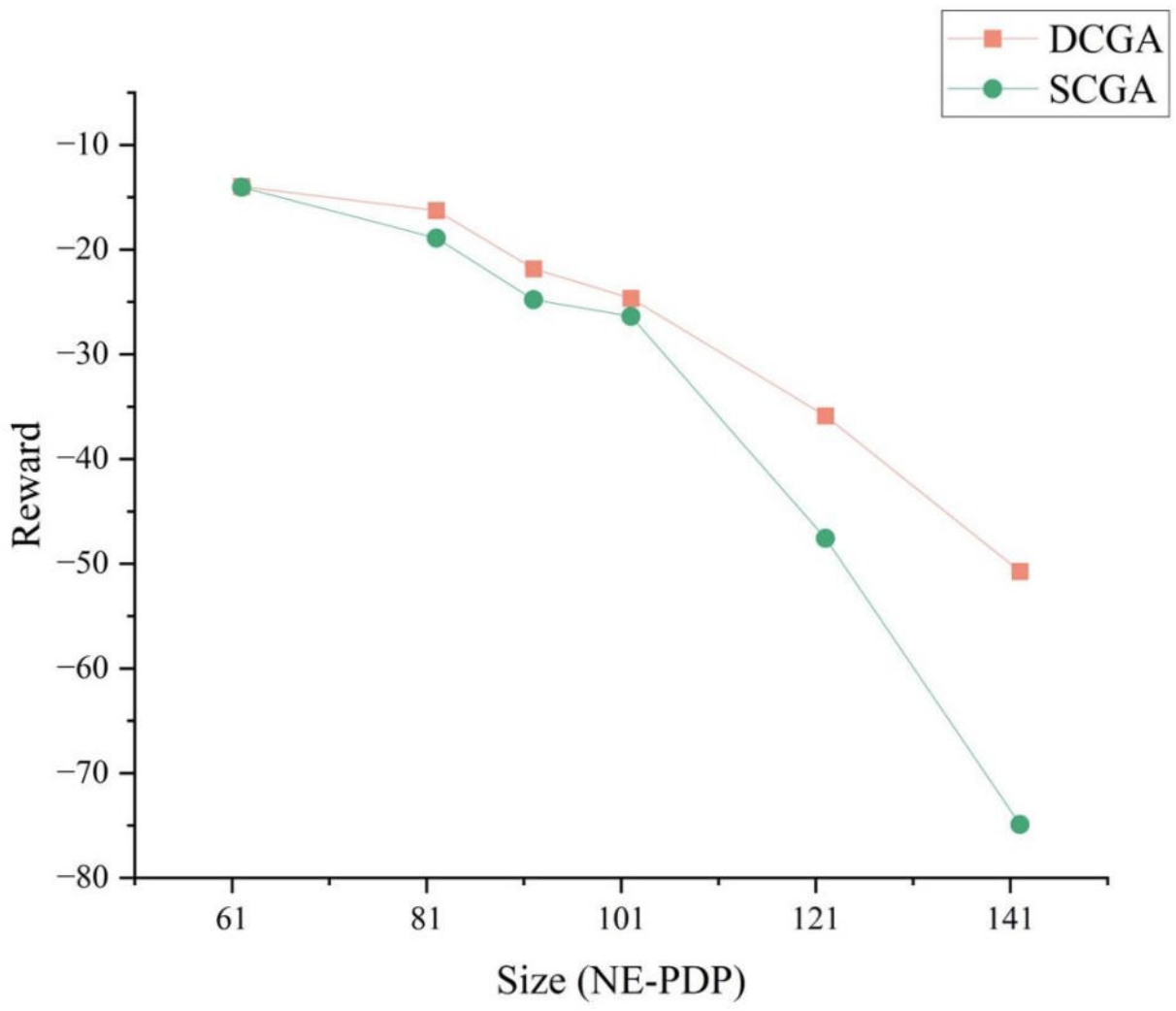

**Figure 9 Single-channel vs. double-channel ablation.**

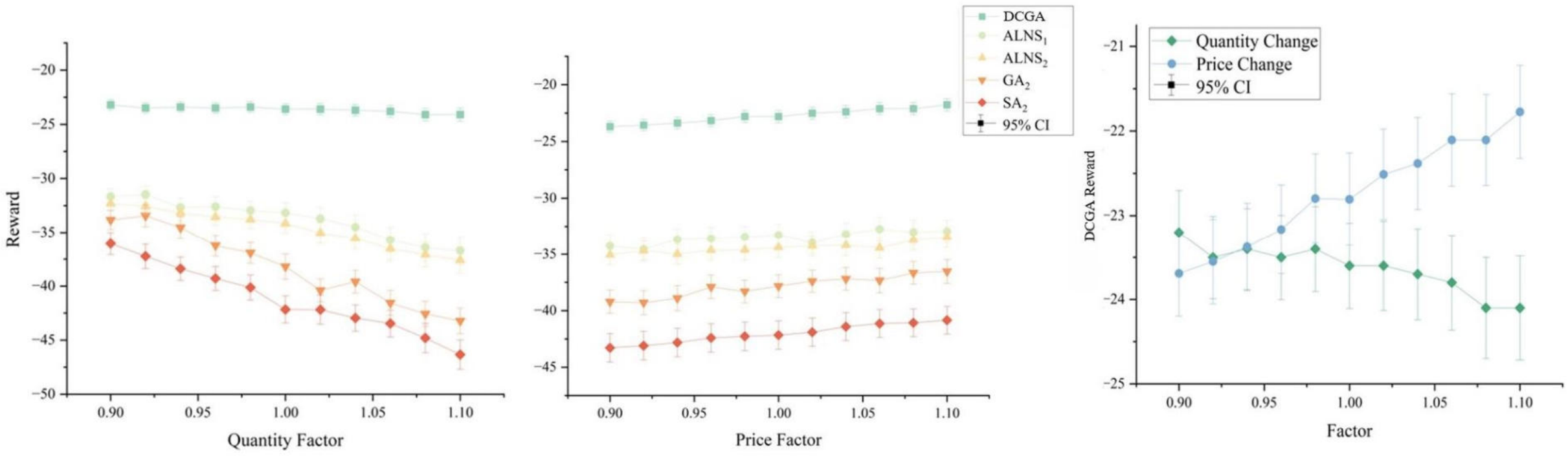

**Figure 10 Model performance under quantity and price perturbations.**

To test this stability, we keep the network structure and pretrained weights fixed, and perturb the realized demand parameters by multiplying each request quantity or price by a scalar factor. Because these perturbations have negligible impact on computational time, we report only solution quality for the quality-oriented benchmark variants. As shown in Fig. 10, DCGA maintains relatively stable performance across a range of quantity and price perturbations. Although its objective value changes as demand parameters vary, the degradation is substantially smaller than that of the benchmark methods, suggesting that the learned policy remains robust under demand shifts beyond the training conditions.

### 4.6. Sensitivity Analysis

To identify which architectural choices most strongly affect learning and final solution quality, we conduct a sensitivity analysis on key hyperparameters. Appendix I of the Online Supplement presents the detailed experimental results.

## 5. Conclusions and Future Work

This paper investigates the NE-PDP-CCS in sparse, directed networks. In this highly constrained setting, cyclic routing is deeply coupled with cargo flow allocation. We first formulate the problem as a rigorous step-indexed MINLP model that preserves continuous operational time. This formulation explicitly captures the mapping between logical and physical nodes, alongside capacity-coupled flow distributions. Subsequently, we develop DCGA, an end-to-end reinforcement learning framework. DCGA seamlessly integrates a double-channel graph representation and GAT-Transformer encoding with a state-conditioned decoder, ultimately utilizing constraint-aware policy shaping and simulator-coupled reward evaluation to construct valid, high-quality solutions.

Computational experiments on the classic maritime dataset, LinerLib, demonstrate that as instance sizes scale, DCGA consistently achieves the best objective values among all tested methods, while maintaining seconds-level inference speeds. Its advantage becomes more pronounced as instance size increases, indicating that structure-aware learning is particularly useful when sparse reachability and routing-allocation coupling make iterative search difficult. Ablation studies indicate that the double-channel representation and feasibility-aware masks are important contributors to performance, while stability and sensitivity analyses suggest that the learned policy is robust within the tested perturbation ranges and that graph feature extraction is a key modeling component.

Several directions deserve further investigation. First, extending the framework from a single service loop to multi-route planning would broaden its applicability to system scheduling, shared capacity, and coordinated transshipment. Second, richer stochastic and dynamic elements, such as time-dependent travel time, congestion, disruptions, and demand uncertainty, could be incorporated through scenario-aware training, rolling re-planning, and robust reinforcement learning. Finally, combining DCGA with lightweight improvement search and developing theoretical characterizations of feasibility and approximation behavior on sparse graphs are promising steps toward connecting learning-based routing policies with classical optimization guarantees.

## Acknowledgments

This research was partially supported by the National Natural Science Foundation of China (Grant Nos. 72371141 and 72301305). The authors would like to thank Hanwen Dai and Tong Zhao for their assistance with data preparation and the development of an experimental environment consistent with practical operational settings. The authors are also grateful to Hanlin Zheng for his valuable support in model validation.

## Appendix A: Constrained Markov Decision Process Formulation and Bellman Equation

The route-construction process can be viewed as a finite-horizon constrained Markov decision process (CMDP). We denote it by

$$\mathcal{M} = (\mathcal{S}, \mathcal{A}, P, r, C, \mathbf{d}, \mu_0, \bar{T}), \tag{S.1}$$

where $\mathcal{S}$ is the state space, $\mathcal{A}$ is the action space, $P$ is the transition kernel, $r$ is the reward function, $C = \{c_i^{con}\}_{i=1}^m$ denotes the constraint cost functions, $\mathbf{d} = (d_1, \dots, d_m)$ denotes the corresponding admissible limits, $\mu_0$ is the initial-state distribution, and $\bar{T}$ is the finite decision horizon.

***State Space $\mathcal{S}$.*** As shown in Fig. 2, the state $s_t \in \mathcal{S}$ contains both static information and dynamic information. The neural representation of these state components is described in Sec. 3.2.

***Action Space $\mathcal{A}$ and Constraints.*** An action $a_t \in \mathcal{A}$ corresponds to selecting the next logical node to visit. Given state $s_t$, only actions in the state-dependent feasible set $\mathcal{A}_F(s_t) \subseteq \mathcal{A}$ are allowed:

$$\pi(a \mid s_t) = 0, \qquad \forall a \notin \mathcal{A}_F(s_t). \tag{S.2}$$

The feasible set $\mathcal{A}_F(s_t)$ incorporates the operational restrictions that can be verified before an action is taken, including sparse-arc feasibility, capacity availability, visit restrictions, termination conditions, and the structural restrictions defined in Sec. 2. In implementation, this support restriction is imposed through the masking mechanism described in Sec. 3.2.

For completeness, these operational requirements can also be expressed in the standard CMDP form. For each constraint $c_i^{\text{con}} \in C$, let

$$J_{c_i^{\text{con}}}(\pi) = \mathbb{E}_{\tau \sim \pi, P}\left[\sum_{t=0}^{\bar{T}-1} c_i^{\text{con}}(s_t, a_t, s_{t+1})\right], \qquad i = 1, \dots, m \tag{S.3}$$

denote its expected cumulative violation, with admissible limit $d_i$. For example, if $B_{t+1}$ denotes the accumulated route duration and $C_{\max}$ the maximum allowable duration, the duration constraint can be represented by $c_{\text{dur}}^{\text{con}}(s_t, a_t, s_{t+1}) = [B_{t+1} - C_{\max}]_+$ and $d_{\text{dur}} = 0$, where $[x]_+ = \max\{x, 0\}$. The admissible policy class is defined as

$$\Pi_C = \left\{\pi \in \Pi : J_{c_i^{\text{con}}}(\pi) \le d_i,\ i = 1, \dots, m, \quad \pi(a \mid s) = 0,\ \forall a \notin \mathcal{A}_F(s)\right\}. \tag{S.4}$$

***State Transition $P$.*** The transition logic is deterministic. Given the current state $s_t$ and the selected action $a_t$, the system transitions to $s_{t+1}$ by updating the vehicle's location, accumulated travel time, residual capacity, onboard load, and service status.

***Reward $r$.*** The objective is evaluated after a complete route has been generated. Let $\tau = (s_0, a_0, \dots, a_{\bar{T}-1}, s_{\bar{T}})$ denote a decoded trajectory and let $R(\tau)$ be the terminal objective value defined by Eq. (2). We use a sparse episodic reward:

$$r_{t+1}(s_t, a_t, s_{t+1}) = \begin{cases} 0, & t+1 < \bar{T}, \\ R(\tau), & t+1 = \bar{T}. \end{cases} \tag{S.5}$$

Although the objective is evaluated only after route completion, the sequential construction still admits the standard value-function interpretation over the feasible action set. For a policy $\pi$ supported on $\mathcal{A}_F(s_t)$, the value and action-value functions are

$$V^\pi(s_t) = \mathbb{E}_\pi[R(\tau) \mid s_t], \qquad Q^\pi(s_t, a_t) = \mathbb{E}_\pi[R(\tau) \mid s_t, a_t]. \tag{S.6}$$

Here the stage index is omitted because the remaining construction information is included in the state representation. With discount factor $\gamma = 1$, they satisfy

$$V^{\pi}(s_t) = \mathbb{E}_{a_t \sim \pi(\cdot|s_t)} [Q^{\pi}(s_t, a_t)], \quad Q^{\pi}(s_t, a_t) = \mathbb{E}[r_{t+1}(s_t, a_t, s_{t+1}) + V^{\pi}(s_{t+1}) \mid s_t, a_t], \quad a_t \in \mathcal{A}_F(s_t). \tag{S.7}$$

The corresponding optimality equations over feasible actions are

$$V^*(s_t) = \max_{a_t \in \mathcal{A}_F(s_t)} Q^*(s_t, a_t), \tag{S.8}$$

$$Q^*(s_t, a_t) = \mathbb{E}\left[ r_{t+1}(s_t, a_t, s_{t+1}) + \max_{a_{t+1} \in \mathcal{A}_F(s_{t+1})} Q^*(s_{t+1}, a_{t+1}) \mid s_t, a_t \right]. \tag{S.9}$$

## Appendix B: Illustrative Example of LPL Projection

Fig. S.1 illustrates how LPL projection aggregates demand-side relations when multiple logical service nodes map to the same physical node. In this example, physical node $N_2$ corresponds to the multi-task set $\{O_2, D_1\}$. Before aggregation, $O_2$ receives a logical relation from $O_1$, while $D_1$ has outgoing logical relations to $D_2$ and $D_3$. After logical-physical projection, these incident relations are merged at $N_2$. Applying the same operation to all multi-task physical nodes transforms the original one-to-one logical demand graph into the aggregated graph shown on the right.

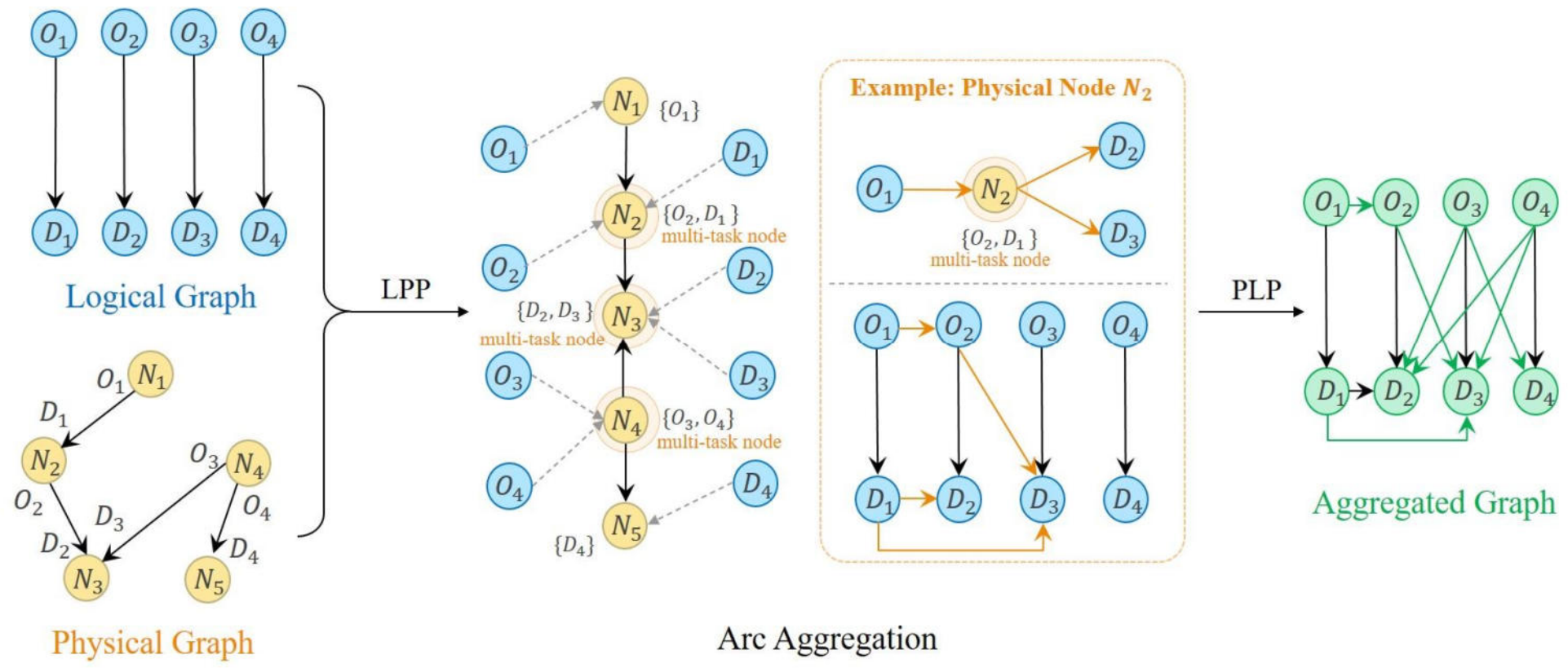

**Figure S.1 Example of LPL projection: multiple logical service nodes mapped to the same physical nodes.**

## Appendix C: Smoothed Boltzmann Distribution under Entropy Regularization

LEMMA 1 **(Boltzmann Form of an Entropy-Regularized Statewise Surrogate)**. *Fix a decoding step $t$, trajectory, the realized state $(s_t)$, and the episode-level advantage $A_i$. For brevity, let*

$$\mathcal{A}_t := \mathcal{A}_F(s_t), \qquad \pi_{\theta,t}(a) := \pi_\theta(a \mid s_t), \tag{S.10}$$

*where $\mathcal{A}_t$ is the nonempty feasible action set retained by the hard masks, and assume $\pi_{\theta,t}(a) > 0$ for all $a \in \mathcal{A}_t$. Holding $\pi_{\theta,t}$ fixed, introduce an auxiliary distribution $q_t \in \Delta_t$, where*

$$\Delta_t := \left\{ q_t : q_t(a) \ge 0, \ \sum_{a \in \mathcal{A}_t} q_t(a) = 1 \right\}. \tag{S.11}$$

*For $\beta > 0$ and $u_t(a) := \alpha A_i \log \pi_{\theta,t}(a)$, the auxiliary problem*

$$\max_{q_t \in \Delta_t} \sum_{a \in \mathcal{A}_t} q_t(a) u_t(a) + \beta \mathcal{H}(q_t), \qquad \mathcal{H}(q_t) := -\sum_{a \in \mathcal{A}_t} q_t(a) \log q_t(a), \tag{S.12}$$

*has the unique maximizer*

$$q_t^\star(a) = \frac{\exp\big(u_t(a)/\beta\big)}{\sum_{a'\in\mathcal{A}_t}\exp\big(u_t(a')/\beta\big)}, \qquad a\in\mathcal{A}_t. \tag{S.13}$$

*Equivalently, we can let*

$$q_t^\star(a) = \frac{\pi_{\theta,t}(a)^{\alpha A_i/\beta}}{\sum_{a'\in\mathcal{A}_t}\pi_{\theta,t}(a')^{\alpha A_i/\beta}}, \qquad a\in\mathcal{A}_t.$$

Proof. The Lagrangian of (S.12) is

$$\mathcal{L}(q_t,\lambda) = \sum_{a\in\mathcal{A}_t} q_t(a)u_t(a) - \beta\sum_{a\in\mathcal{A}_t} q_t(a)\log q_t(a) + \lambda\left(\sum_{a\in\mathcal{A}_t} q_t(a) - 1\right). \tag{S.14}$$

At an interior stationary point,

$$\frac{\partial\mathcal{L}}{\partial q_t(a)} = u_t(a) - \beta\big(1+\log q_t^\star(a)\big) + \lambda = 0, \tag{S.15}$$

$$q_t^\star(a) = C\exp\big(u_t(a)/\beta\big) = C\exp\left(\frac{\alpha A_i \log\pi_{\theta,t}(a)}{\beta}\right). \tag{S.16}$$

Normalization gives (S.13). The objective in (S.12) is strictly concave because its first term is linear in $q_t$ and $\beta\mathcal{H}(q_t)$ is strictly concave for $\beta>0$. Thus, the stationary point is the unique global maximizer. Moreover, $\pi_{\theta,t}(a)>0$ implies that all scores are finite, so $q_t^\star(a)>0$ for every $a\in\mathcal{A}_t$. This auxiliary problem provides a statewise variational interpretation; it is not an independent statewise optimization performed by the parametric actor. □

## Appendix D: Decoder State-Conditioning Details

As shown in Fig. S.2, this appendix details the two state-conditioning modules in the decoder. Let $H^{(N_2)}\in\mathbb{R}^{B\times|\mathcal{V}|\times D}$ be the static node tensor produced by the encoder, where $B$ is the batch size and $D$ is the hidden dimension.

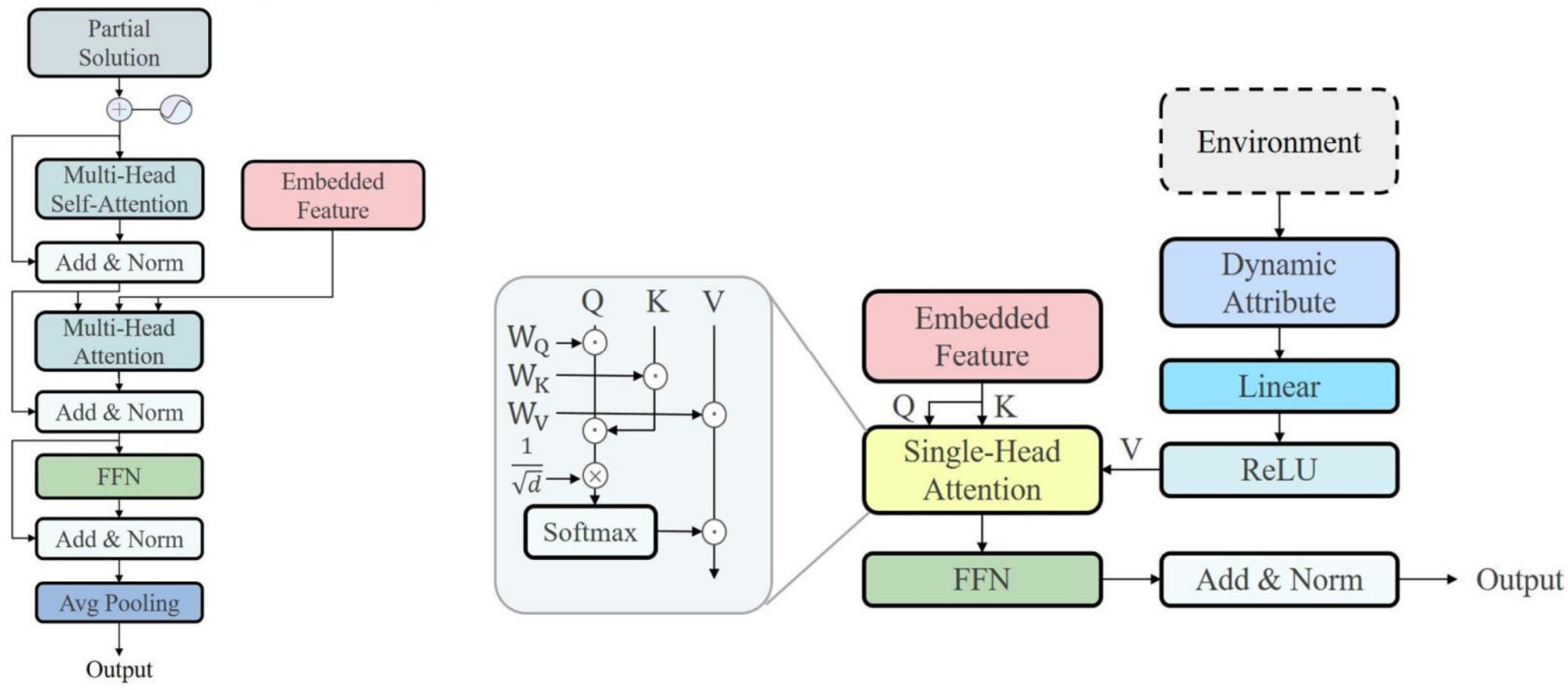

**Figure S.2** **The micro-encoder (left) summarizes the partial route, while the conditional attention gate (right) injects dynamic time-capacity information into candidate-node representations.**

***Micro encoder.*** The micro-encoder captures the order information in the partial route $p_t=(a_0,\dots,a_{t-1})$. It first extracts the corresponding node embeddings from $H^{(N_2)}$, then applies positional encoding and a lightweight Transformer encoder:

$$S_t = H^{(N_2)}[:,p_t,:], \qquad \tilde{S}_t = \mathrm{PE}(S_t), \qquad S_t^{\mathrm{seq}} = \mathrm{TF}(\tilde{S}_t). \tag{S.17}$$

The partial-route representation is obtained by mean pooling,

$$h_t^{\mathrm{seq}} = \frac{1}{|p_t|}\sum_{\ell=1}^{|p_t|} S_{t,\ell}^{\mathrm{seq}}, \tag{S.18}$$

with a trainable start-history vector used when $p_t$ is empty.

***Conditional attention gate.*** The conditional attention gate injects dynamic simulator information into the candidate-node representations without re-running the graph encoder. We collect the remaining-time and residual-capacity signals as $\Xi_t = [x_t^{\text{time}}; x_t^{\text{cap}}]$, and obtain a dynamic-conditioned node tensor:

$$\hat{H}_t = \text{CondGate}_\omega(H^{(N_2)}, \Xi_t), \tag{S.19}$$

where $\omega$ denotes the trainable parameters of the conditional gate.

Finally, the decoder query combines the graph-level summary ($\bar{h}$), current-node ($c_t$) embedding, and partial-route representation ($h_t^{\text{seq}}$):

$$z_t = [\bar{h};\ H^{(N_2)}[:, c_t, :];\ h_t^{\text{seq}}], \tag{S.20}$$

and the state-conditioned context is computed by

$$h_t^c = \text{MHA}(z_t, \hat{H}_t, \hat{H}_t). \tag{S.21}$$

Substituting $h_t^c$ into Eq. (37) gives the raw scores used by the policy-shaping module.

## Appendix E: Critic Details

Fig. S.3 shows the critic architecture.

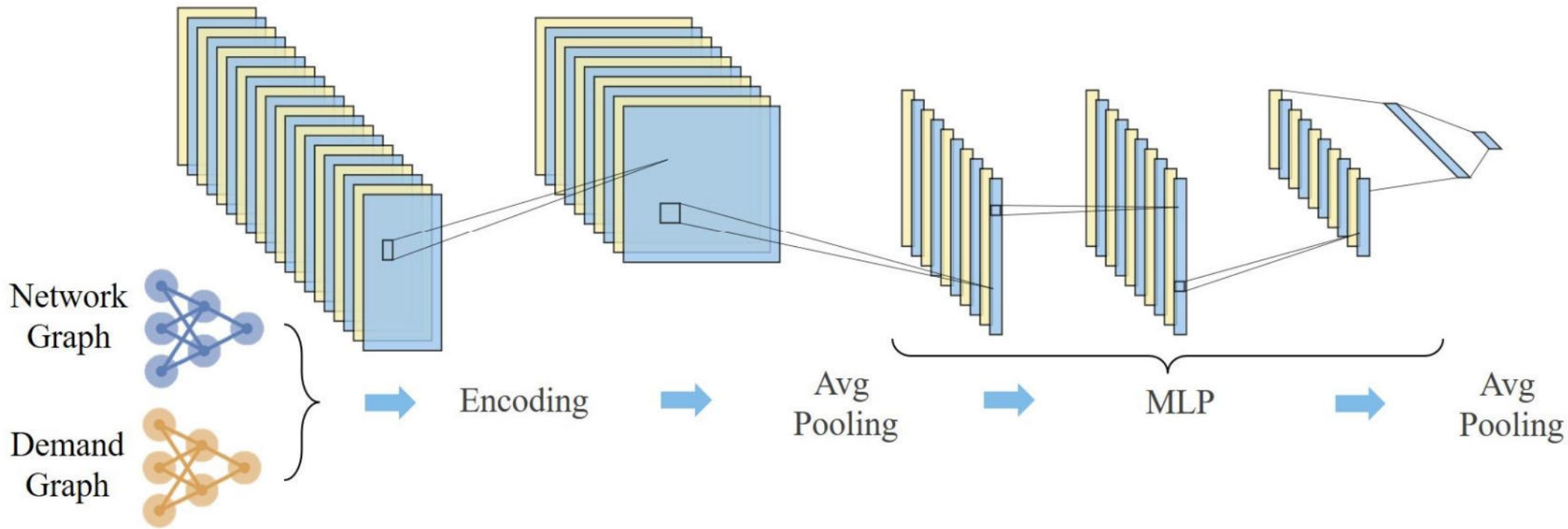

**Figure S.3 Critic network architecture.**

The double-channel GAT streams and the Transformer refinement produce the static node feature and global context as in Sec. 3.2.1 and Sec. 3.2.2:

$$H^{(N_2)'} \in \mathbb{R}^{B\times(2N+1)\times D}, \qquad \bar{h}' = \frac{1}{2N+1}\sum_{i\in\mathcal{V}} H^{(N_2)'}_{:,i,:} \in \mathbb{R}^{B\times D}. \tag{S.22}$$

Let the initial state be $s_0$, which is embedded as $x^{\text{crit}} = \bar{h}'$ by the critic encoder. A three-layer multilayer perceptron (MLP) decoder maps $x^{\text{crit}}$ to a scalar value estimate $V_\psi(H^{(N_2)'})$ per batch element:

$$\begin{cases} h^{(1)} = \sigma\big(x^{\text{crit}} W_1 + b_1\big), \\ h^{(2)} = \sigma\big(h^{(1)} W_2 + b_2\big), \\ V_\psi(H^{(N_2)'}) = h^{(2)} W_3 + b_3, \quad V_\psi(H^{(N_2)'}) \in \mathbb{R}^{B\times 1}. \end{cases} \tag{S.23}$$

## Appendix F: Implementation Details and Benchmark Settings

All computational experiments were conducted on a high-performance workstation equipped with an Intel(R) Xeon(R) Platinum 8581C CPU and an NVIDIA RTX PRO 5000 GPU. Table S.1 summarizes the key architectural and optimization hyperparameters used in training and inference. Table S.2 reports the main parameter settings for all compared baselines.

**Table S.1 Implementation details and training hyperparameters of DCGA.**

| Item | Setting |
|---|---|
| Graph Encoder | layers = 4, heads = 8, embedding dim = 128 |
| Fusion Encoder | layers = 4, heads = 8, embedding dim = 128, FFN dim = 256 |
| Actor Decoder | heads = 8, embedding dim = 128 |
| Delivery bias | 5 |
| Tanh coef. | 100 |
| Advantage coef. | $\alpha = 1$ |
| Entropy coef. | $\beta = 0.01$ |
| Actor Optimizer | Adam, learning rate = $1 \times 10^{-4}$ |
| Critic Optimizer | Adam, learning rate = $5 \times 10^{-5}$ |
| Clipping | gradient norm clip = 1.0 |
| Batching | 2 batches per episode |
| Hardware | Intel(R) Xeon(R) Platinum 8581C CPU and an NVIDIA RTX PRO 5000 GPU |

**Table S.2 Major benchmark configuration parameters.**

| Method | Key settings |
|---|---|
| MIP (Gurobi) | time limit = 3600s; warm start = greedy; threads = auto |
| Flow Model (Gurobi) | time limit = no limit |
| ALNS | depth = 10, width = 10, retry = 3, max single change = 3 |
| GA | population = 50; generations = 100; elitism size = 5; tournament size = 3; mutation rate = 0.3 |
| SA | $T_0$ = 5000; cooling = 0.995; steps = 1500 |
| LKH3 | time limit = 2 h / instance, max trials = 5000, runs = 10 |
| NCS | Inference steps = 5 + #nodes, sample size = 64, others keep consistent with Kong et al. (2024). |

Table S.3 lists the five neighborhood operators used in our ALNS-based polishing procedure. Table S.4 presents the rules for scoring based on the assessment results. Operator selection is sampled according to a weight-proportional distribution, and the weights are adapted by an exponential moving average (EMA) using the average score of each operator within a short window.

**Table S.3 ALNS neighborhood operators.**

| Operator | Description |
|---|---|
| Relocate | Remove one node at position $i$ and reinsert it at a different position $j$. |
| Swap | Randomly pick two positions and exchange their nodes. |
| Segment reverse | Select a segment $[i, j]$ and reverse the order of nodes in the segment. |
| Add port | Append a new port not currently in the subsequence, limited by the extend ratio. |
| Remove node | Delete a randomly chosen node from the subsequence, preserving a minimum length ratio. |

**Table S.4 ALNS controller scoring mechanism design.**

| Status | Score | Meaning |
|---|---|---|
| Best | 5.0 | Produces a new global best solution |
| Improved | 2.0 | Improves over the current solution |
| Accepted | 0.5 | Accepted feasible solution |
| Rejected | 0.0 | Rejected infeasible solution |

## Appendix G: Adaptation Details of Benchmarks

***McCormick-relaxed optimization benchmark.*** The original formulation contains the bilinear soft due-date commitment penalty $w_r\delta_r$, where both the fulfilled quantity $w_r$ and tardiness $\delta_r$ are continuous variables. To obtain a mixed-integer linear optimization benchmark solvable by Gurobi, we replace $g_r = w_r\delta_r$ with its McCormick relaxation over $0 \le w_r \le q_r$ and $0 \le \delta_r \le C_{\max}$:

$$0 \le g_r \le q_r\delta_r, \qquad g_r \le C_{\max}w_r, \qquad g_r \ge q_r\delta_r + C_{\max}w_r - q_rC_{\max}, \quad \forall r \in \mathcal{R}. \tag{S.24}$$

Since both variables are continuous, this construction is a relaxation rather than an exact linearization. For any feasible solution of the original bilinear formulation, setting $g_r = w_r\delta_r$ satisfies the above inequalities, so the original feasible region is contained in the relaxed feasible region. Moreover, because the objective maximizes revenue minus the penalty term $\sum_{r\in\mathcal{R}} \varphi_r g_r$, the relaxation can only reduce the represented penalty and hence yields an objective value no smaller than that of the original bilinear model:

$$Z^{\text{rel}}_{\text{MILP}} \ge Z^{\text{orig}}. \tag{S.25}$$

Therefore, the relaxed MILP should be interpreted as an optimistic optimization benchmark, or an upper-bound reference, rather than an exact certificate for the original bilinear problem. This treatment strengthens the comparison in favor of the optimization benchmark: DCGA still outperforms it on medium- and large-scale instances despite the relaxed MILP being allowed to overestimate the true objective value.

***LKH3 adaptation.*** LKH3 (Available at `http://webhotel4.ruc.dk/~keld/research/LKH-3/`) (Helsgaun 2017) is a highly effective heuristic framework for TSP and several vehicle-routing variants. Since it does not directly support our full routing-and-flow problem, we use the closest available variant, the multi-commodity pickup-and-delivery traveling salesman problem (m-PDTSP), as the routing core. Each OD request is treated as one commodity, and LKH3 generates a physical-port sequence. Table S.5 summarizes the main mismatches between LKH3 and our problem and the corresponding adaptations. Thus, LKH3 should be interpreted as a strong but structurally imperfect routing benchmark.

**Table S.5 Adaptation of LKH3 to the NE-PDP-CCS.**

| Mismatch | Adaptation |
|---|---|
| Sparse network | Assign a large penalty ($10^7$) to infeasible arcs. |
| Composite objective and cargo flow allocation | Use LKH3 only to generate the routing sequence; evaluate the final cargo flow allocation under the original objective using linear programming. |
| Maximum duration, cyclic service, optional fulfillment, and capacity | Treat all OD requests as necessary commodities in LKH3, and allow partial or unmet fulfillment in the subsequent cargo flow allocation stage. |

***NCS adaptation.*** NCS (Available at `https://github.com/yining043/PDP-N2S`) (Kong et al. 2024) is the closest available learning-based reference for PDP, but its original representation is designed for Euclidean planar instances. We therefore construct two adapted variants, summarized in Table S.6. Both variants are trained with the same batch size and the same two-batch-per-epoch protocol as DCGA, while the remaining hyperparameters follow the original NCS. They are trained until convergence, defined as no improvement on the validation set for 20 consecutive evaluation rounds.

**Table S.6 Adaptation of NCS to the NE-PDP-CCS.**

| Variant | Construction |
|---|---|
| **NCS**$_1$ | Uses port longitude and latitude as coordinate inputs and augments the decoder with a feasibility mask to enforce basic sparse-network connectivity. |
| **NCS**$_2$ | Adds the same graph encoder used in DCGA before the NCS backbone, so that sparse reachability and arc attributes can be injected while retaining the original NCS decoder. Other DCGA-specific mechanisms, such as additional feasibility masks, are not incorporated. |

## Appendix H: Pseudocode for the Training Algorithm

**Algorithm 1** Entropy-Regularized Monte Carlo Policy Gradient Training

**Require:** Training dataset $\mathcal{D}$, batch size $B$, learning rates $\eta_\theta, \eta_\psi$, epochs $E$

1: Initialize actor parameters $\theta$ and critic parameters $\psi$
2: **for** epoch = 1 to $E$ **do**
3:     **for** batch $b = 1$ to $|\mathcal{D}|/B$ **do**
4:         Sample a batch of $B$ instances $\{s_0^{(i)}\}_{i=1}^B$ from $\mathcal{D}$
5:         Compute graph embeddings $H_i^{(N_2)}$ for all $i \in \{1, \dots, B\}$
6:         **Parallel Decoding:**
7:         **for** $t = 0$ to $\bar{T} - 1$ **do**
8:             Sample actions $a_t^{(i)} \sim \pi_\theta(\cdot \mid s_t^{(i)})$ for all $i$
9:             Update states $s_{t+1}^{(i)} \leftarrow \text{EnvStep}(s_t^{(i)}, a_t^{(i)})$
10:             Store log-probabilities $\log p_t^{(i)} \leftarrow \log \pi_\theta(a_t^{(i)} \mid s_t^{(i)})$
11:         **end for**
12:         Calculate episodic return $R^{(i)} = \sum_{t=0}^{\bar{T}-1} r_t^{(i)}$ for all $i$
13:         Estimate baseline $b^{(i)} \leftarrow V_\psi(H_i^{(N_2)})$
14:         Compute advantages $A^{(i)} \leftarrow R^{(i)} - b^{(i)}$
15:         **Parameter Update:**
16:         $\mathcal{L}_\pi \leftarrow -\frac{1}{B} \sum_{i=1}^B (\alpha A^{(i)} \sum_{t=0}^{\bar{T}-1} \log p_t^{(i)} + \beta \sum_{t=0}^{\bar{T}-1} \mathcal{H}(\pi_\theta(\cdot \mid s_t^{(i)}))$.
17:         $\mathcal{L}_v \leftarrow \frac{1}{2B} \sum_{i=1}^B (b^{(i)} - R^{(i)})^2$
18:         $\theta \leftarrow \text{Adam}(\theta, \nabla_\theta \mathcal{L}_\pi, \eta_\theta)$
19:         $\psi \leftarrow \text{Adam}(\psi, \nabla_\psi \mathcal{L}_v, \eta_\psi)$
20:     **end for**
21: **end for**

## Appendix I: Sensitivity Analysis of Key Hyperparameters

The model consists of (i) a *Graph Encoder* that extracts graph-structured representations from the sparse non-Euclidean networks, and (ii) a *Fusion Encoder* that further integrates multi-source features before decoding. We vary one hyper-parameter at a time while keeping the training protocol, data, and all other settings fixed, and report the corresponding learning curves in Fig. S.4. The results suggest that the overall performance might be primarily constrained by *graph feature extraction quality*.

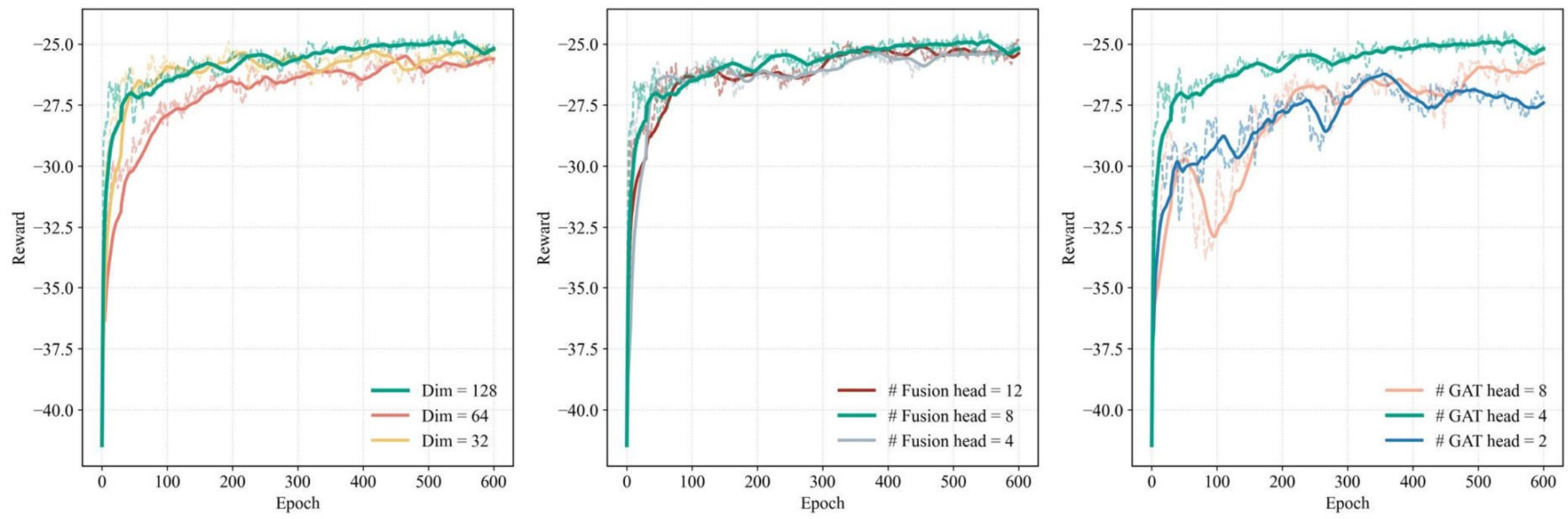

**Figure S.4 Sensitivity analysis of key hyperparameters.**

***Embedding dimension.*** Increasing the encoder embedding dimension consistently improves both learning speed and asymptotic performance (Fig. S.4, left). A larger embedding space enables richer instance representations, which is particularly beneficial for NE-PDP where routing decisions must reflect long-range dependencies and routing-allocation coupling. However, the performance gain exhibits diminishing returns at larger dimensions, suggesting a practical trade-off between solution quality and model size.

***Fusion Encoder heads.*** Varying the number of attention heads in the Fusion Encoder has only a marginal impact on the learning trajectory and final reward (Fig. S.4, middle). This indicates that the fusion stage mainly plays a stabilizing integration role, and its multi-head configuration is not a bottleneck once a reasonable capacity is provided.

***Graph Encoder heads.*** In contrast, the Graph Encoder is notably sensitive to the number of attention heads (Fig. S.4, right). Too few heads restrict the diversity of relational patterns captured from sparse graph neighborhoods, whereas too many heads may dilute head-wise specialization and introduce optimization noise, both leading to inferior convergence.